%% file: main.tex
\documentclass[10pt,twocolumn,letterpaper]{article}

\usepackage[pagenumbers]{cvpr}   

\input{config}

\input{acronyms}
\input{notations}
\input{content/macros.tex}

\begin{document}

    \input{title}
    
    \begin{abstract}
    \input{content/abstract}
    \end{abstract}
    
    \input{content/introduction.tex}
    \input{content/related_work.tex}

    \input{content/method.tex}
    \input{content/experiments.tex}

    \input{content/conclusion.tex}

    {\small
    \bibliographystyle{ieee_fullname}
    \bibliography{content/references.bib}
    }
    
    \clearpage
    \include{content/supplementary_material}

\end{document}

%% file: config.tex
\usepackage{graphicx}
\usepackage{amsmath}
\usepackage{amssymb}
\usepackage{booktabs}

\usepackage[pagebackref,breaklinks,colorlinks]{hyperref}

\usepackage[capitalize]{cleveref}
\crefname{section}{Sec.}{Secs.}
\Crefname{section}{Section}{Sections}
\Crefname{table}{Table}{Tables}
\crefname{table}{Tab.}{Tabs.}

\def\cvprPaperID{6524} 
\def\confName{CVPR}
\def\confYear{2023}

\usepackage{hyperref}
\hypersetup{colorlinks=true, linkcolor=blue, citecolor=blue}
\usepackage{multirow}
\usepackage{booktabs}
\usepackage{amsmath}
\usepackage[dvipsnames]{xcolor}
\usepackage{xspace}
\usepackage{xcolor}
\usepackage{siunitx}
\usepackage{MnSymbol}  
\usepackage{xstring}   
\usepackage{gensymb}   
\usepackage{mfirstuc}  
\usepackage{hhline}    
\usepackage{pifont}    
\usepackage{enumitem}  
\usepackage{tabularx}  
\usepackage{contour}   
\usepackage{ulem}      

\contourlength{0.8pt}

\newcommand{\myuline}[1]{%
  \uline{\phantom{#1}}%
  \llap{\contour{white}{#1}}%
}

\makeatletter
\newcommand\notsotiny{\@setfontsize\notsotiny{6.8}{7.5}}
\makeatother

\newcommand{\cmark}{\textcolor{OliveGreen}{\ding{51}}}%
\newcommand{\xmark}{\textcolor{BrickRed}{\ding{55}}}%
\newcommand{\negmark}{\textcolor{BrickRed}{\textbf{\textminus}}}%
\newcommand{\beginsupplement}{       
        \numberwithin{figure}{section}
        \numberwithin{table}{section}
     }

\newrobustcmd\customBold{\DeclareFontSeriesDefault[rm]{bf}{b}\bfseries}                          %

\let\origparagraph\paragraph
\renewcommand{\paragraph}[1]{\vspace{-3mm}\origparagraph{#1}}

\newcommand{\blockcomment}[1]{}
\newcommand{\biovil}{BioViL\xspace}
\newcommand{\temporalbiovil}{\mbox{BioViL-T}\xspace}  

%% file: acronyms.tex
\usepackage[nolist]{acronym}

\makeatletter
\newcommand*{\org@overidelabel}{}
\let\org@overridelabel\@verridelabel
\@ifpackagelater{acronym}{2015/03/21}{
  \renewcommand*{\@verridelabel}[1]{%
    \@bsphack
    \protected@write\@auxout{}{\string\AC@undonewlabel{#1@cref}}%
    \org@overridelabel{#1}%
    \@esphack
  }%
}{
  \renewcommand*{\@verridelabel}[1]{%
    \@bsphack
    \protected@write\@auxout{}{\string\undonewlabel{#1@cref}}%
    \org@overridelabel{#1}%
    \@esphack
  }%
}
\makeatother

\begin{acronym}
    \acro{AR}{auto-regressive}
    \acro{CNN}{convolutional neural network}
    \acro{CXR}{chest X-ray}
    \acro{FOV}{field of view}
    \acro{MLM}{masked language modelling}
    \acro{MHSA}{multi-head self-attention}
    \acro{MI}{mutual information}
    \acro{NEM}{named entity metric}
    \acro{NLI}{natural language inference}
    \acro{NN}{nearest-neighbour}
    \acro{SOTA}{state-of-the-art}
    \acro{SSL}{self-supervised learning}
    \acro{TEM}{temporal entity matching}
    \acro{ViT}{vision transformer}
    \acro{VLP}{vision--language processing}
\end{acronym}

%% file: notations.tex
\newcommand{\real}{\mathbb{R}}

\newcommand{\imgembdim}{D_{\mathrm{img}}}
\newcommand{\txtembdim}{D_{\mathrm{txt}}}
\newcommand{\jointembdim}{D}
\newcommand{\npatches}{L}
\newcommand{\wimg}{W}
\newcommand{\himg}{H}
\newcommand{\nimgs}{T}
\newcommand{\wpatches}{W'}
\newcommand{\hpatches}{H'}
\newcommand{\nmhsalayers}{K}
\newcommand{\numtexttokens}{M}
\newcommand{\numscans}{Z}
\newcommand{\batch}{\mathcal{B}}

\newcommand{\curr}{\mathrm{curr}}
\newcommand{\prev}{\mathrm{prior}}
\newcommand{\sample}{\mathbf{x}}

\newcommand{\currimg}{\mathbf{x}_\mathrm{img}^\curr}
\newcommand{\previmg}{\mathbf{x}_\mathrm{img}^\prev}
\newcommand{\currreport}{\mathbf{x}_\mathrm{txt}^\curr}

\newcommand{\report}{\mathbf{x}_\mathrm{txt}}

\newcommand{\phrase}{\mathbf{w}}
\newcommand{\currphrase}{\mathbf{w}_\mathrm{txt}^\curr}
\newcommand{\prevphrase}{\mathbf{w}_\mathrm{txt}^\prev}

\newcommand{\patchtokens}{\mathbf{P}}
\newcommand{\patchtoken}{\mathbf{p}}

\newcommand{\currtokens}{\patchtokens^\curr}

\newcommand{\vitfeats}{\mathbf{H}}
\newcommand{\currfeats}[1]{\vitfeats^\curr_{#1}}
\newcommand{\prevfeats}[1]{\vitfeats^\prev_{#1}}

\newcommand{\clsfeats}{\mathbf{r}}
\newcommand{\clsprojfeats}{\mathbf{r^\mathrm{proj}}}
\newcommand{\bertprojection}{\phi_\mathrm{txt}}
\newcommand{\imgprojection}{\phi_\mathrm{img}}

\newcommand{\difftokens}{\patchtokens^\mathrm{diff}}

\newcommand{\missingtoken}{\patchtoken^\mathrm{miss}}

\newcommand{\visualtokens}{\mathbf{V}}
\newcommand{\visualtoken}{\mathbf{v}}

\newcommand{\projvisualtoken}{\mathbf{v}^\mathrm{proj}}
\newcommand{\meanprojvisualtoken}{\bar{\mathbf{v}}^\mathrm{proj}}

\newcommand{\spatialenc}{\mathbf{S}}
\newcommand{\temporalenc}{\mathbf{T}}
\newcommand{\currenc}{\mathbf{t}^\curr}
\newcommand{\prevenc}{\mathbf{t}^\prev}

\newcommand{\clstoken}{\texttt{[CLS]}}
\newcommand{\mlmtoken}{\texttt{[MLM]}}
\newcommand{\septoken}{\texttt{[SEP]}}
\newcommand{\masktoken}{\texttt{[MASK]}}

\newcommand{\imgencoder}{E_\mathrm{img}}
\newcommand{\txtencoder}{E_\mathrm{txt}}
\newcommand{\transformer}{A_\mathrm{img}}

\newcommand{\layernorm}{\operatorname{LN}}
\newcommand{\concat}[2]{#1 \oplus #2}  

\newcommand{\singleimageset}{\mathcal{D}_s}
\newcommand{\multiimageset}{\mathcal{D}_m}
\newcommand{\alldataset}{\mathcal{D}}
\newcommand{\testset}{\mathcal{D^\mathrm{test}}}

\newcommand{\Findings}{\textsc{Findings}}
\newcommand{\Impression}{\textsc{Impression}}

\newcommand{\MLMloss}{\mathcal{L}_\mathrm{MLM}}

\DeclareMathOperator*{\argmax}{\arg\max}

%% file: content/macros.tex
\newcommand{\sourcecodeurl}{\url{https://aka.ms/biovil-t-code}}
\newcommand{\temporaldataurl}{\url{https://aka.ms/ms-cxr-t}}
\newcommand{\modelurl}{\url{https://aka.ms/biovil-t-model}}

\newcommand{\cxrtbenchmark}{\mbox{\textit{MS-CXR-T}}}
\newcommand{\mscxrbenchmark}{\mbox{\textit{MS-CXR}}}

%% file: content/abstract.tex
Self-supervised learning in \ac{VLP} exploits semantic alignment between imaging and text modalities.
Prior work in biomedical \ac{VLP} has mostly relied on the alignment of single image and report pairs even though clinical notes commonly refer to prior images.
This does not only introduce poor alignment between the modalities but also a missed opportunity to exploit rich self-supervision through existing temporal content in the data.
In this work, we explicitly account for prior images and reports when available during both training and fine-tuning.
Our approach, named \temporalbiovil, uses a \acs{CNN}--Transformer hybrid multi-image encoder trained jointly with a text model.
It is designed to be versatile to arising challenges such as pose variations and missing input images across time.
The resulting model excels on downstream tasks both in single- and multi-image setups, achieving \ac{SOTA} performance on (I) progression classification, (II) phrase grounding, and (III) report generation, whilst offering consistent improvements on disease classification and sentence-similarity tasks.
We release a novel multi-modal temporal benchmark dataset, \cxrtbenchmark{}, to quantify the quality of vision--language representations in terms of temporal semantics.
Our experimental results show the advantages of incorporating prior images and reports to make most use of the data.

\acresetall

%% file: content/introduction.tex
\section{Introduction}
\label{sec:intro}

Self-supervision from image--text pairs has enabled the development of flexible general-purpose vision--language models both in the general domain~\cite{radford2021learning,li2021align,yu2022coca} and for specialised domains such as biomedicine and radiology \cite{zhang2020contrastive,huang2021gloria,boecking2022making}.
\Ac{VLP} has shown that cross-modal supervision can provide a richer signal for training both image~\cite{desai2021virtex} and text~\cite{boecking2022making} models. However, the success of \ac{VLP} relies on paired samples sharing semantics, i.e., given an image and text pair, the text should describe the image with minimal extraneous detail~\cite{chuang2020debiased,jia2021scaling,chuang2022robust}. 

\input{content/figure_tex/fig_overview}

In this regard, \ac{VLP} in biomedicine and radiology poses a distinctive challenge, as reports routinely include comparisons to prior imaging studies~\cite{american2020acr,aideyan1995influence,rousan2020chest}. Without knowledge of this prior image\footnote{In the MIMIC-CXR v2 dataset \cite{johnson2019mimic}, around \emph{40\% of reports} explicitly reference a previous image. See \cref{sec:mimic_appendix_longitudinal} for details.}, temporal information in the text modality, e.g. ``Pneumonia is improving'', could pertain to any image containing ``Pneumonia'', producing ambiguity during contrastive training (\Cref{figure:motivation}).
Despite this, the existing \ac{VLP} work to date considers alignment between only single images and reports~\cite{zhang2020contrastive,huang2021gloria,moon2022multi,boecking2022making}, going so far as to remove temporal content from reports in training data to prevent `hallucinations' in downstream report generation\cite{ramesh2022improving}.
However, temporal information can provide complementary self-supervision, solely by exploiting existing structure, and without requiring any additional data.

In this work, we neither ignore nor remove temporal information in the text modality, but explicitly account for it during pre-training. 
Rather than treating all image--report pairs in the dataset as independent, we exploit temporal correlations by making prior images available for comparison to a given report.
To learn from this structure, we develop a temporal \ac{VLP} pre-training framework named  \textit{\temporalbiovil{}}. A core component is its new multi-image encoder that can handle the absence of prior images and potential spatial misalignment between images across time.
\temporalbiovil{} takes into account prior images where available, removing cross-modal ambiguity as illustrated in \cref{figure:motivation}. Linking multiple images during pre-training proves beneficial to both image and text models: we report \ac{SOTA} performance on both temporal image classification and report generation. In the latter case, we show that prefixing the prior \emph{report} substantially increases performance, again reflecting the value of prior information.
We emphasise that the benefit is not restricted to temporal downstream tasks: our approach also achieves \ac{SOTA} on non-temporal tasks of pneumonia detection \cite{shih2019augmenting} and phrase grounding \cite{ms-cxr-benchmark}, underscoring the value of a cleaner learning signal during \ac{VLP} without  needing to modify or add to the training dataset.
Our contributions can be summarised as follows:
\begin{itemize}[parsep=1pt, topsep=0pt, itemsep=1pt, leftmargin=5mm]
    \item We introduce a novel pre-training framework called \textit{\temporalbiovil{}}. It leverages the temporal relationship of samples to self-supervise \ac{VLP} models, making commonly used biomedical \ac{VLP} models (e.g., \cite{boecking2022making, huang2021gloria, zhang2020contrastive}) more applicable to a wider range of downstream tasks without compromising performance on existing benchmarks.
    \item We develop a generic multi-image encoder that handles missing image inputs and incorporates longitudinal information without requiring explicit image registration.
    \item We achieve \ac{SOTA} results in \ac{CXR} report generation, temporal image classification, and phrase grounding downstream benchmarks by accounting for prior context in self-supervised training and fine-tuning.
    \item We release a new multimodal benchmark dataset, \cxrtbenchmark{}, curated by an expert radiologist. It enables benchmarking of \ac{CXR} \ac{VLP} models in terms of temporal semantics extracted from image and text data.
\end{itemize}

%% file: content/figure_tex/fig_overview.tex
\begin{figure}[t!]
    \centering
    \includegraphics[width=\linewidth]{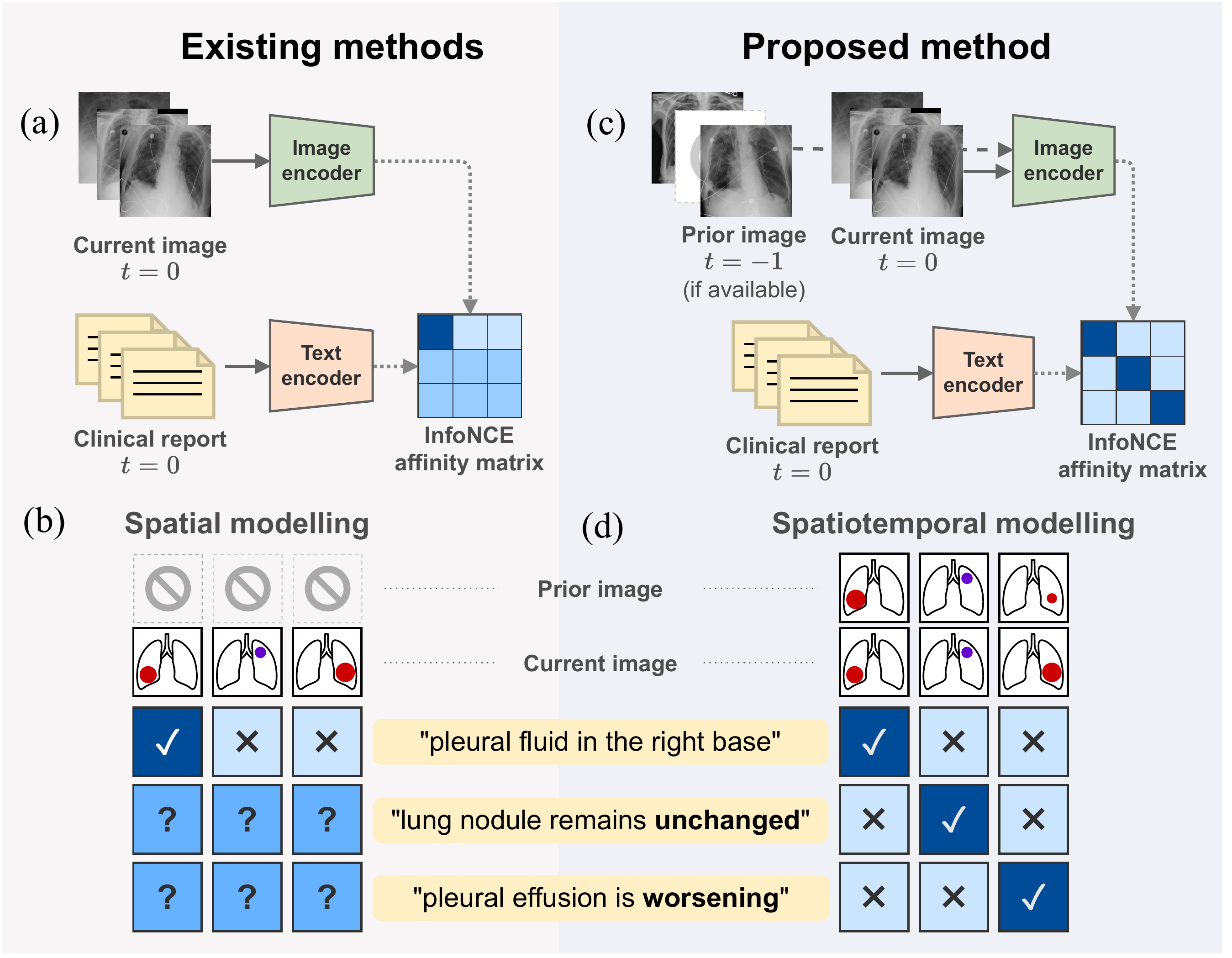}
    \caption{
        (a) Existing visual--language pre-training approaches \cite{boecking2022making, huang2021gloria, zhang2020contrastive} often use only a single image for contrastive learning (e.g., InfoNCE \cite{oord2018representation}).
        (b) In such settings, discarding the temporal connectivity of images limits the alignment of image--text pairs as shown with the affinity matrix, leading to suboptimal pre-training and missed opportunity to create additional model supervision for free.
        (c, d) Our approach exploits this domain knowledge by learning to incorporate a series of images and correlate them to reports, leading to pre-trained models that can generalise to a wider range of downstream tasks whilst achieving \acs{SOTA} performance.
    }
    \label{figure:motivation}
\end{figure}

%% file: content/related_work.tex
\section{Related work}
\vspace{3mm} 

\paragraph{Vision--language processing}
Self-supervised \ac{VLP} can significantly reduce the need for manual labels required for the training of image encoders\cite{radford2021learning,desai2021virtex}. The availability of large-scale paired image--text datasets has thus led to rapid development of general-purpose \ac{VLP} models. Objectives include contrastive and discriminative image--text matching \cite{radford2021learning,li2021align,wang2021vlmo} including local variants~\cite{huang2021gloria,yao2021filip}, \ac{AR} captioning~\cite{yu2022coca,li2022blip,alayrac2022flamingo} and multi-modal masked modelling objectives~\cite{chen2020uniter,li2021align,Singh2022FLAVAAF}.

\paragraph{Biomedical vision--language processing} 
Paired medical image--report datasets were originally used for supervised learning via (typically) automated label extraction from clinical reports\cite{wang2017chestx,irvin2019chexpert,smit2020chexbert}. Using such datasets, advances in general-domain \emph{self-supervised} \ac{VLP} have been demonstrated to benefit biomedical imaging applications\cite{zhang2020contrastive,huang2021gloria,boecking2022making}. Work has incorporated ideas from general-domain \ac{VLP} such as the original CLIP-style cross-modal contrastive objective~\cite{zhang2020contrastive}, multi-modal masking with merged co-attention on image--text representations~\cite{moon2022multi}, and adaptations to the data of the domain. For example, a radiology report may have sparse image-specific details, prompting a local modification to the contrastive loss enabling alignment between text tokens and image patches~\cite{huang2021gloria}. Domain-specific pre-training of the text model is shown to benefit biomedical \ac{VLP}~\cite{boecking2022making}, and preferential masking of medical terms during \ac{MLM} was explored~\cite{yan2022clinical}. Here we use a local loss and domain-specific pre-training of the text model, but did not find a benefit to preferential masking. Similarly, cross-attention~\cite{dou2022empirical} is used  rather than merged co-attention for image-guided \ac{MLM}.

\paragraph{Longitudinal modelling of medical images}
While prior images are used in unimodal \emph{supervised} longitudinal analysis of medical images
\cite{karwande2022chexrelnet,Santeramo2018LongitudinalDO,Wang2019TowardsPT,Xu2019DeepLP}, temporal information has not directly been employed for self-supervision. The closest work exploits patient metadata to select positive or negative examples in unimodal contrastive learning~\cite{vu2021medaug,zeng2021contrastive}.

Existing models typically employ either late fusion of global image representations \cite{Santeramo2018LongitudinalDO,Wang2019TowardsPT,Xu2019DeepLP,Sriram2021COVID19PV}, which can miss fine-grained localised changes \cite{huang2021gloria}, or explicit spatial correspondence of features, using fixed spatial grids \cite{oh2019longitudinal} or object detection \cite{karwande2022chexrelnet}.
Registering image pairs is commonly used for change detection in other contexts \cite{daudt_fully_2018,peng_end--end_2019,shi_change_2020}, and has been applied to medical imaging\cite{avants2008diffeomorphic,durrleman2013spatiotemporal}. For \acp{CXR} however, registration entails the ill-posed problem of aligning 2D projections of 3D geometry, 
which inevitably results in residual misalignment. Our approach does not rely on bounding boxes or explicit graph construction as it uses self-attention of visual tokens across time to handle any spatial misalignment.

\paragraph{Self-supervision across time}
Self-supervision has found applications on densely-sampled time series data (e.g., video) to capture temporal information~\cite{han2020self,recasens2021broaden,zeng2021contrastiveB,yun2022time}. Our problem setting involves sparsely and sporadically sampled data where temporal pretext tasks are less applicable\cite{agrawal2022leveraging}. Similarly, it requires text supervision to enable both static and temporal learning, when temporal structure is present.

%% file: content/method.tex
\begin{figure*}[t!]
    \centering
    \includegraphics[width=0.9\textwidth,trim={18mm 0 0 5mm},clip]{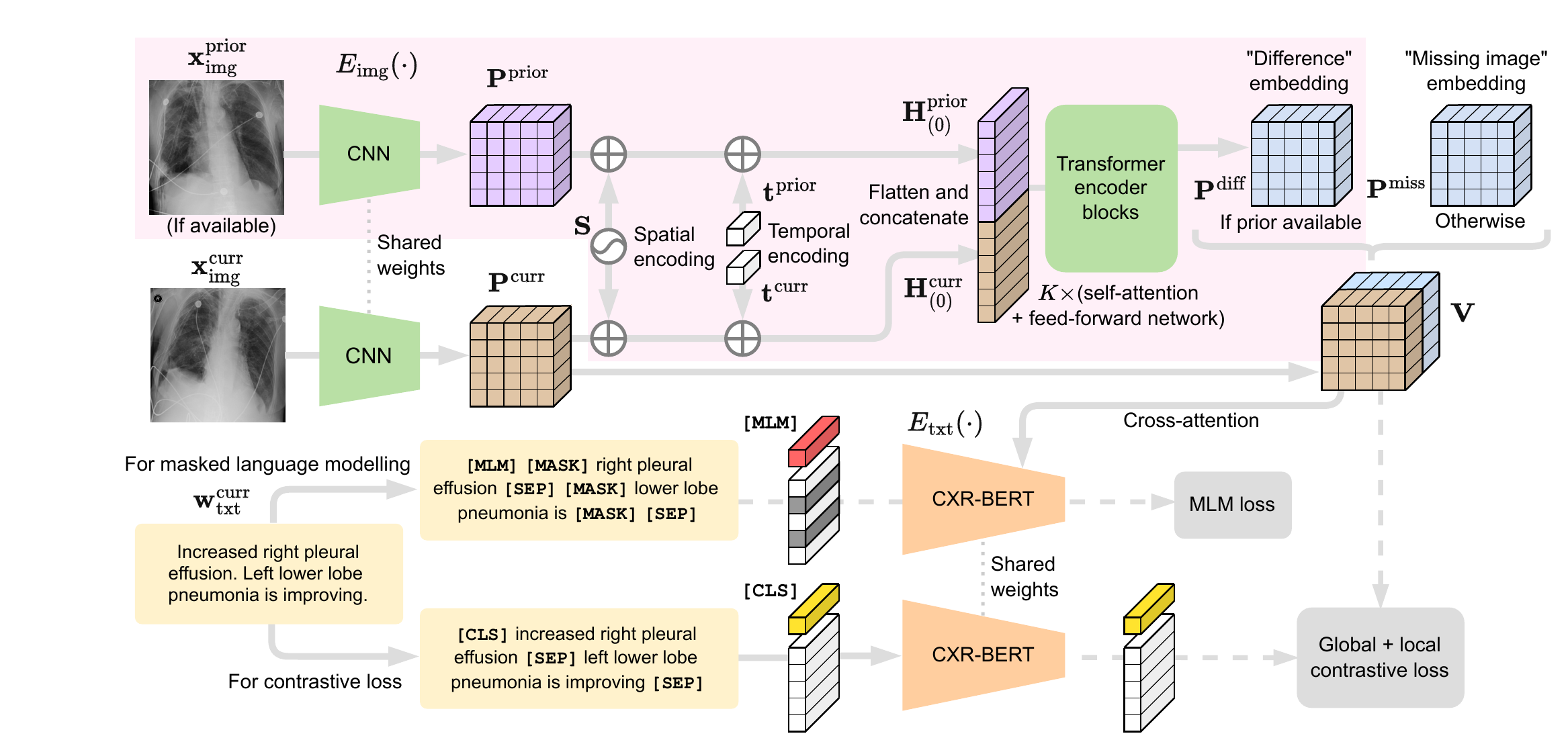}
    \caption{
        The proposed self-supervised \ac{VLP} training framework \temporalbiovil:
        Image representations $\visualtokens$ are extracted from single and multiple input scans (whenever available) using a hybrid CNN and transformer encoder \cite{park2022how, d2021convit}. This design choice is to increase the data-efficiency and enable the fusion of temporal content without requiring image registration.
        They are later matched with their corresponding text representations obtained with CXR-BERT \cite{boecking2022making} using local \cite{huang2021gloria} and global InfoNCE \cite{oord2018representation} training objectives.
        As an additional model supervision, multi-modal fused representations, obtained with cross-attention, are used for image-guided masked language modelling.
    }
    \label{fig:biovilt_diagram}
\end{figure*}

\section{BioViL-T training framework}
Our approach comprises a multi-image encoder designed to extract spatio-temporal features from sequences of images (\Cref{sec:image_features}) and a text encoder incorporating optional cross-attention on image features. The models are trained jointly with image-guided \ac{MLM} and cross-modal global and local contrastive objectives (\Cref{sec:training_objectives}). The resulting image and text models are later adapted for uni- or multi-modal downstream tasks as described in \Cref{sec:adaptations_to_downstream}. Implementation details are presented in \Cref{sec:implementation_details,sec:experiment_details}.

For a given image and report pair  $(\currimg, \currreport)$, the report $\currreport$ describes the current image content and changes in reference to prior images. Our proposed formulation focuses on a single prior image; however, it can be generalised to \emph{multiple} prior images depending on the application. Hence, we construct datasets by including the prior image whenever it exists\footnote{The prior \emph{report} is not included during pre-training as it may further reference an earlier study, reintroducing temporal ambiguity.}: $(\currimg, \previmg, \currreport) \in \multiimageset$ or $(\currimg, \varnothing, \currreport) \in \singleimageset$ with the resulting dataset being a union of single and multi-image examples: $\alldataset = \multiimageset \cup \singleimageset$.

\subsection{Extracting spatio-temporal image features}
\label{sec:image_features}
Clinical findings are often observed across different image regions and co-occur simultaneously, which requires dense level visual reasoning across time to capture both static and temporal features. In contrast to late global fusion \cite{Sriram2021COVID19PV} and bounding-box based approaches \cite{karwande2022chexrelnet}, \temporalbiovil\ leverages local correspondences between image regions across time using transformer self-attention blocks \cite{dosovitskiy2020image}. Thus our method does not require an explicit image registration step between time points.

We propose a hybrid CNN--Transformer encoder model due to its data efficiency and spatial flexibility of cross-attention across time points: $\imgencoder: \real^{\wimg \times \himg} \rightarrow \real^{\wpatches \times \hpatches \times \imgembdim}$ (e.g., ResNet-50 \cite{he2016resnet}) and $\transformer: \real^{\nimgs \times \npatches \times \imgembdim} \rightarrow \real^{\npatches \times \imgembdim}$ (e.g., transformer \cite{dosovitskiy2020image}), where $\wimg$, $\himg$, and $\nimgs$ correspond to spatiotemporal dimensions, $\npatches=\wpatches\hpatches$ is the number of visual tokens per image, and  $\imgembdim$ is the embedding dimension. 
Here $\imgencoder$ serves as a stem network \cite{park2022how} to provide visual token features of individual images. The CNN's inductive biases \cite{park2022how, d2021convit} ensure data efficiency of our hybrid model, making it ideal for smaller scale biomedical datasets. $\imgencoder$ is initialised with \biovil\ weights~\cite{boecking2022making}. The main purpose of $\transformer$ is to capture patch embedding interactions across time when a prior image $\previmg$ is available and to aggregate them into a fixed-length token representation. Input visual tokens, $\currfeats{0} = \currtokens := \imgencoder(\currimg)$, $\prevfeats{0} := \imgencoder(\previmg)$ are augmented with spatio-temporal positional encodings and flattened across the spatial dimensions. They are then processed by $\nmhsalayers$ transformer encoder \cite{vaswani2017attention} layers $A$ as follows:
\begin{equation}
    \begin{bmatrix}\currfeats{k} \\ \prevfeats{k} \end{bmatrix} = A_k\!\left(\begin{bmatrix}
        \currfeats{k-1} + \spatialenc + \mathbf{1}_\npatches\otimes\currenc \\
        \prevfeats{k-1} + \spatialenc + \mathbf{1}_\npatches\otimes\prevenc
    \end{bmatrix}\right) ,
\end{equation}
for $k=1,\dots,\nmhsalayers$, where $\spatialenc \in \real^{\npatches\times\imgembdim}$ denotes 2D sinusoidal positional encodings \cite{carion2020end} and $\temporalenc = [\currenc; \prevenc] \in \real^{2\times\imgembdim}$ is its temporal counterpart, which is learnt (\cref{fig:biovilt_diagram})~\cite{alayrac2022flamingo}.
The layer-normalised ($\layernorm$) \cite{ba2016layernorm} output of the final transformer encoder block ${\difftokens := \layernorm(\currfeats{\nmhsalayers})}$ is an `aggregated' representation of patch-level progression information anchored on the current image. \Cref{fig:self_attention_pose_variations} shows attention roll-out \cite{abnar-zuidema-2020-quantifying} applied to $\difftokens$ after pre-training, showing how the prior image contributes to the fused representation. \Cref{fig:self_attention_pose_variations_supp}  further highlights the robustness to variations in pose underlining that registration is not necessary for this encoder.

\paragraph{Static-temporal feature decomposition}
When a prior image is available the final image representation $\visualtokens := \concat{\currtokens}{\difftokens} \in \real^{\wpatches \times \hpatches \times 2\imgembdim} $ is formed by concatenating two sets of features (similar to \cite{behrmann2021long}): those from the current image alone ($\currtokens$) and the temporal features from current and prior images ($\difftokens$). In this way, self-attention is mainly required to cope with pose variations and patch comparisons across time in extracting temporal content, removing the need for registration or explicit spatial feature alignment. 
When no prior scan is available ($\sample \in \singleimageset$), $\transformer$ is not used and $\difftokens$ is replaced by a learnable token $\missingtoken \in \real^{\imgembdim}$, replicated across the spatial dimensions. \Cref{sec:disentangled_features_ablation} later demonstrates that $\transformer$ highlights the value of feature decomposition for tasks such as phrase grounding which require well-localised features \cite{ms-cxr-benchmark}.

Hereafter, downstream tasks that require solely single image features, $\currtokens$, are referred to as \textit{static tasks}, and the ones that benefit from additional progression information, $\difftokens$, as \textit{temporal tasks}, e.g., report decoding. 

\subsection{Text-supervision for spatio-temporal learning}
\label{sec:training_objectives}
Let $\phrase=(w_1,\dots,w_\numtexttokens)$ denote a vector of $\numtexttokens$ tokens of a report $\report$ after tokenisation. We first obtain contextualised token features $\txtencoder(\phrase) \in \real^{\numtexttokens \times \txtembdim}$ by passing a sequence of text tokens $\phrase=(w_1,\dots,w_\numtexttokens)$ through a BERT encoder $\txtencoder$ \cite{devlin2018bert}. The input sequence is prepended with either a $\clstoken$ or $\mlmtoken$ token associated with a downstream training objective, conditioning the output features similar to \cite{li2022blip, liu2018generating}. During training, we do two forward passes through $\txtencoder$: once with masking at 45\% probability (for the \ac{MLM} objective) and once without masking for contrastive learning, as shown in \Cref{fig:biovilt_diagram}. The text encoder is initialised with the weights of CXR-BERT\footnote{\url{https://huggingface.co/microsoft/BiomedVLP-CXR-BERT-general}} \cite{boecking2022making} canonical model, trained on domain-specific vocabulary and corpora.

Both text and image features are later projected into a joint latent space with $\bertprojection: \real^{\txtembdim} \rightarrow \real^\jointembdim$, and similarly $\projvisualtoken_{w, h} := \imgprojection (\visualtoken_{w, h})$ where $\imgprojection : \real^{\imgembdim} \rightarrow \real^{\jointembdim}$, with $\phi$ being a two-layer perceptron in our experiments. 

\paragraph{Contrastive objectives} Let $\clsfeats := [\txtencoder(\phrase)]_{\clstoken}$ denote the global representation of $\phrase$, with $\clsprojfeats := \bertprojection(\clsfeats)$ its projected version. Given projected patch embeddings $\projvisualtoken_{w, h}$, we can compute a global cosine similarity $S_{C}(\meanprojvisualtoken, \clsprojfeats)$ and a local similarity using weighted pairwise cosine similarities across text tokens and projected patch embeddings~\cite{huang2021gloria,yao2021filip}. These similarities are used in both global and local contrastive objectives with the InfoNCE loss~\cite{oord2018representation,radford2021learning}. The local loss proves crucial both for static phrase-grounding and temporal image classification (see \Cref{tab:ablation_experiments}), highlighting the importance of localised self-supervision.

\paragraph{Image-guided masked language modelling}
Prior work \cite{boecking2022making, moon2022multi} has shown that biomedical visual-language learning benefits from an auxiliary task such as \ac{MLM} since capturing the joint distribution of tokens can stabilise and improve language understanding during joint learning. Given a batch $\batch$ of token vectors $\phrase$, it is often defined as the cross-entropy for predicting the randomly sampled masked tokens, $m \subset \{1,\dots,\numtexttokens\}$, $\MLMloss = -\frac{1}{|\batch|} \sum_{\phrase \in \batch} \log p_\theta(\phrase_m \,|\, \phrase_{\backslash m})$, where $\theta$ are the weights of the text encoder $\txtencoder$. 

In the absence of image information, however, certain masked findings and attributes are not readily predicted, e.g., ``$\masktoken$ is worsening''. As shown in the general domain \cite{chen2020uniter}, visual information can help disambiguate such masked predictions and provide additional cross-modal supervision. Thus, we use cross-attention~\cite{vaswani2017attention,dou2022empirical} to the image features $\projvisualtoken_{w,h}$ during this task. Specifically, for our image-guided \ac{MLM} objective we model $p_\theta(\phrase_m \,|\, \phrase_{\backslash m}, \projvisualtoken_{w, h})$.

\subsection{Adaptations to downstream tasks}
\label{sec:adaptations_to_downstream}
\temporalbiovil\ can be adapted to various downstream tasks. For phrase-grounding and zero-shot inference, we rely on $S_C(\clsprojfeats$, $\projvisualtoken_{w, h})$ similar to \cite{boecking2022making, huang2021gloria}. For multiple-text prompts, projected text embeddings are marginalised prior to $\ell_2$-normalisation \cite{radford2021learning}. To enable language decoding, $\projvisualtoken_{w, h}$ inputs are cross-attended by text queries $\phrase$, and causal-attention is utilised between text tokens \cite{li2022blip, vaswani2017attention}. 
Differing from \cite{boecking2022making, huang2021gloria, zhang2020contrastive}, we show that report generation tasks can greatly benefit from temporal joint latent space. 

\paragraph{Conditioning on prior reports} In contrast to existing work, we incorporate the prior report as a prompt to contextualise the report generation task: $p_{\Phi} (\currphrase | \, \prevphrase , \, \projvisualtoken_{w, h})$, where $\Phi$ are the multi-modal encoder--decoder network's weights, and $\currphrase$, $\prevphrase$ denote text tokens for current and prior reports respectively. This is analogous to fine-tuning GPT-3 \cite{brown2020language} with prompts and instructions \cite{wei2021finetuned}, but conditioning on both images and the previous report. A dedicated separation token $\septoken$ is added into the input sequence $[\prevphrase, \septoken, \currphrase]$.

\newcommand{\imgc}{\mathbf{x}^\mathrm{img}}

\paragraph{Curation of imaging datasets} CXR datasets \cite{johnson2019mimic} often contain multiple image acquisitions $\mathcal{Z} = \{ \imgc_1, \dots, \imgc_{\numscans} \}$ in a single visit due to data quality issues such as a limited field-of-view or scanning the wrong body part (\Cref{fig:data_curation_supp}).
Unlike \cite{boecking2022making, huang2021gloria, zhang2020contrastive}, we conduct curation to choose higher quality images among the potential candidates instead of performing a random selection.
For this step, a separate \temporalbiovil{} is trained on `clean' studies with single acquisitions and later used in a zero-shot setting to detect out-of-distribution samples \cite{esmaeilpour2022zero,fort2021exploring} arising from the re-imaging process.
The candidate $\hat{z}$ is selected as follows: $\hat{z} = \argmax_{z \in \mathcal{Z} } S_{C} ( \meanprojvisualtoken_z \, , \, \clsprojfeats)\;\: \textrm{s.t.} \;\: | s_{\hat{z}} - s_{\mathcal{Z}\backslash \hat{z}} | > \delta$ for a margin $\delta$.
This approach is applied to enhance the quality of the temporal classification dataset given its limited size.

%% file: content/experiments.tex
\section{Datasets \& experiments}
Here, we demonstrate \temporalbiovil's data efficiency and adaptability to a wide range of applications, and show how the model achieves SOTA performance on various downstream tasks by learning from data instances linked across time, making effective use of domain priors and the available training data.
Specifically, our model is evaluated on a diverse set of downstream tasks including zero- and few-shot static and temporal image classification, report generation, phrase-grounding \cite{ms-cxr-benchmark}, and sentence similarity.

\paragraph{\cxrtbenchmark{} benchmark} We release a new multi-modal benchmark dataset\footnote{\cxrtbenchmark{} benchmark dataset can be accessed through PhysioNet: \temporaldataurl{}}, \cxrtbenchmark{}, to evaluate chest X-ray VLP models on two distinct temporal tasks: image classification and sentence similarity. The former comprises multi-image and ground-truth label pairs ($N=1326$) across $5$ findings, with classes corresponding to $3$ states of disease progression for each finding: \{\texttt{Improving}, \texttt{Stable}, \texttt{Worsening}\}. The latter quantifies the temporal-semantic similarity of text embeddings extracted from pairs of sentences ($N=361$). The pairs can be either paraphrases or contradictions in terms of disease progression. The data for both tasks was manually annotated and reviewed by a board certified radiologist. \cref{sec:cxrt-benchmark} provides further details on its data distribution and annotation protocol.

\input{content/figure_tex/fig_attention.tex}

\paragraph{Datasets} For pre-training, we use the MIMIC-CXR v2 \cite{johnson2019mimic, goldberger2000physiobank} chest X-ray dataset, which contains longitudinal imaging studies with corresponding radiological reports, see \cref{fig:histogram_longitudinal_studies} for the distribution of studies. We only use frontal view scans and discard samples where reports do not contain an \Impression\ section. From this data, we gather $174.1$k and $4.9$k text-image pairs for training and validation respectively, with a majority of pairs including a prior image: $|\multiimageset^{\mathrm{train}}|=118.8k$, $|\singleimageset^{\mathrm{train}}|=55.3k$. The text consists of the \Impression{} section and, for \ac{MLM} additionally the \Findings{} section if available. Note that \emph{no manual labels} are used during pre-training and \emph{no additional data} is used for the methods that leverage the link between current and prior images. For early stopping we track the validation loss, see \cref{sec:implementation_details} for implementation details.

Downstream evaluations are performed on a disjoint held-out test set shared across all tasks, $|\testset|=2971$. For report generation, we extend this test set with samples from healthy subjects ($N=815$) to match the prevalence of pathological studies used in prior work~\cite{chen-emnlp-2020-r2gen,miura-etal-2021-improving,endo2021retrieval}. For fine-tuning on temporal image classification, we use labels from the Chest ImaGenome dataset~\cite{chest-imagenome} as in \cite{karwande2022chexrelnet} (statistics in \Cref{table:imagenome_silver}). In detail, we use the following benchmark datasets: (I) \mscxrbenchmark{}~\cite{ms-cxr-benchmark} for phrase grounding, (II) the RSNA Pneumonia dataset~\cite{shih2019augmenting,wang2017chestx} to test zero-shot and fine-tuned classification, (III) \cxrtbenchmark{} for temporal sentence similarity and temporal image classification.

\paragraph{Comparison approaches} We compare our approach to other domain-specific \ac{SOTA} pre-training frameworks \cite{boecking2022making, huang2021gloria} specifically on phrase-grounding and zero-shot predictive performance. The non-temporal BioViL framework~\cite{boecking2022making} is most similar to our approach and provides insight into non-temporal pre-training.
We additionally compare to internal ablations such as removing the past report during report generation and masking prior images during phrase grounding.
For \ac{SOTA} performance comparison, various \ac{AR} and \ac{NN} based language decoding approaches are used as baselines: IFCC~\cite{miura-etal-2021-improving}, R2Gen~\cite{chen-emnlp-2020-r2gen}, CXR-RePaiR-2~\cite{endo2021retrieval}, and CXR-RePaiR-Select \cite{endo2021retrieval}.

For the temporal classification task, we compare against a baseline exploiting the BioViL image encoder~\cite{boecking2022making}, and an approach that makes use of graph convolutions across regions of interest extracted from bounding boxes \cite{karwande2022chexrelnet}. For BioViL, we perform affine image registration (with 4 DoF) for each pair of scans to cope with pose variations, and the encoded images are concatenated along the feature dimension and classified via a multilayer perceptron.  For \cite{karwande2022chexrelnet}, we compare to the three-class setting. Lastly, we benchmark our final text model in isolation against domain specific \ac{SOTA} models in a temporal sentence similarity task: CXR-BERT\cite{boecking2022making} and PubMedBert\cite{gu2021domain}. 

\begin{table}[t!]
    \centering
    \caption{
        Results for report generation task: Predictions are evaluated in terms of lexical (BLEU-4, ROUGE) and factuality metrics (CHEXBERT, TEM). Approaches are grouped into two broad categories: \acf{NN} and \acf{AR}. \temporalbiovil\ pre-training consistently yields improved decoding. Further, the consistent performance gains of using prior image and report demonstrate the importance of such domain priors. `PI~/~PR' indicate usage of prior image and report, respectively.
    }
    \setlength{\tabcolsep}{4pt}
    \resizebox{\linewidth}{!}{
    
    \sisetup{table-format=2.1(1)}
    \begin{tabular}{@{}l@{\hskip 0.2cm}llcSSSS@{}}
        \toprule
        & \textbf{Method}                   & \textbf{Pre-training} & \textbf{PI / PR}  & \textbf{BLEU-4} & \textbf{ROUGE} & \textbf{CHEXBERT} & \textbf{TEM} \\
        \midrule
        \parbox[t]{3mm}{\multirow{3}{*}{\rotatebox[origin=c]{90}{\underline{NN}}}}
        &CXR-RePaiR-2 \cite{endo2021retrieval}      &   \biovil\            &   \xmark\ /  \xmark  &   2.1 &   14.3 &   28.1 & 12.5 \\
        
        &Baseline (NN) \cite{boecking2022making}    &   \biovil\            &   \xmark\ /  \xmark  &  3.7 &   20.0 &   28.3  & 11.1 \\
        
        &Proposed (NN)                              &   \temporalbiovil\    &   \cmark  / \xmark  &   4.5 &   20.5 &   29.0 & 13.0 \\
        
        \midrule
        
        \parbox[t]{3mm}{\multirow{3}{*}{\rotatebox[origin=c]{90}{\underline{AR}}}}
        &Baseline (AR) \cite{boecking2022making}    & \biovil\        & \xmark\ / \xmark & 7.5 \pm 0.1 &   27.9 \pm 0.1 &  29.3 \pm 0.3 & 13.8 \pm 0.1 \\  
        &Proposed                                   & \temporalbiovil & \cmark / \xmark & 8.2 \pm 0.1 &   28.7 \pm 0.1 &  30.2 \pm 0.7  & 16.0 \pm 0.3 \\
        &Proposed                                   & \temporalbiovil & \cmark / \cmark & \customBold 9.2 \pm 0.3 & \customBold 29.6 \pm 0.1 &   \customBold 31.7 \pm 1.0  & \customBold 17.5 \pm 0.1 \\
        \bottomrule
    \end{tabular}}
    \label{table:temporal_decoding_experiments}
\end{table}

\begin{table}[tb]
    \centering
    \caption{Temporal image classification results (repeated for $4$ random seeds) on the \cxrtbenchmark{} benchmark for fully-supervised and zero-/few-shot (Z\&F) learning settings, in terms of macro-accuracy across the three classes for each finding. Affine registration is performed for the baseline method (denoted with suffix `w/reg'), to partially address the pose variations across scans.}
    \setlength{\tabcolsep}{2pt}
    \resizebox{\linewidth}{!}{
    \sisetup{table-format=2.1(1)}
    \begin{tabular}{@{}l@{\hspace{10pt}}llSSSSS@{}}
        \toprule
        & \textbf{Method (\% of labels)}                                      & \textbf{Pre-train} & \textbf{Consolidation} & \textbf{Pl. effusion} & \textbf{Pneumonia} & \textbf{Pneumothorax} & \textbf{Edema} \\
        \midrule
        \parbox[t]{2mm}{\multirow{2}{*}{\rotatebox[origin=c]{90}{\myuline{\strut Z\&F}}}}
        &\temporalbiovil prompt\ (0\%)               & Temporal  & 53.6 \pm 1.9 & 59.7 \pm 2.1 & 58.0 \pm 3.9 & 34.9 \pm 1.0 & 64.2 \pm 1.5 \\
        &\temporalbiovil\ (10\%)                     & Temporal  & 59.7 \pm 2.4 & 62.4 \pm 1.4 & 60.1 \pm 2.1 & 35.3 \pm 2.6 & 62.6 \pm 1.7 \\
        \midrule
        \parbox[t]{2mm}{\multirow{6}{*}{\rotatebox[origin=c]{90}{\myuline{\strut Supervised}}}}
        &CNN + Transformer                              & ImageNet    & 44.0 \pm 2.0 & 61.3 \pm 1.6 & 45.1 \pm 3.5 & 31.5 \pm 3.1 & 65.5 \pm 1.1 \\ 
        &CheXRelNet \cite{karwande2022chexrelnet}    & ImageNet    & 47           & 47 & 47 & 36 & 49                                              \\ 
        &\biovil{}~\cite{boecking2022making}          & Static     & 56.1 \pm 1.5  & 62.3 \pm 1.1  & 59.4 \pm 1.0   & 41.7 \pm 2.8  & 67.5 \pm 0.8 \\
        &\biovil{} w/reg \cite{boecking2022making}    & Static     & 56.0 \pm 1.5  & 63.0 \pm 0.9  & 60.2 \pm 0.7   & 42.5 \pm 2.7  & 67.5 \pm 0.9 \\
        & \temporalbiovil\ wout curation           & Temporal   & 58.9 \pm 1.7 & 65.5 \pm 0.7 & 61.5 \pm 2.2 &  44.4 \pm 2.1  & 67.4 \pm 0.8 \\
        & \temporalbiovil{}                       & Temporal   & \customBold 61.1 \pm 2.4 & \customBold 67.0 \pm 0.8 & \customBold 61.9 \pm 1.9 & 42.6 \pm 1.6 & \customBold 68.5 \pm 0.8 \\
        \bottomrule
    \end{tabular}}
    \label{table:temporal_progression_general}
\end{table}

\paragraph{Metrics} Due to class imbalance, we report macro-accuracy for temporal image classification. For phrase grounding, we use mean Intersection-Over-Union (mIoU) and Contrast-to-Noise-Ratio (CNR) \cite{boecking2022making}. The latter measures the discrepancies between cosine similarities inside and out of the bounding box region without requiring hard thresholds. To evaluate the quality of generated reports, we use both the standard lexical metrics, e.g., BLEU\cite{papineni-etal-2002-bleu}, ROUGE-L\cite{lin-2004-rouge}, and also domain-specific factuality metric: CheXbert\footnote{The average of the weighted-$F_1$ score across 14 pathological observations labelled by CheXbert.} \cite{smit-etal-2020-combining}. To directly probe the generation of change-related information, we introduce a new metric called \ac{TEM} to compute the match score of a fixed set of temporal entities (see \cref{sec:tem_metric}).


\subsection{Temporal pre-training yields data efficiency}
\label{sec:zero_shot_temporal}
\noindent\fbox{\begin{minipage}{23em}
\emph{Downstream tasks are enabled with minimal labels.}
\end{minipage}}\vspace{2mm}

The sections `NN' and `Z\&F' on \Cref{table:temporal_decoding_experiments,table:temporal_progression_general} report zero- and few-shot performance on tasks benefitting from temporal information: temporal image classification and report generation. Here we measure the quality of the learnt joint latent space and the extent to which \temporalbiovil enables efficient use of raw data.
For zero-shot classification we prompt the \ac{AR} fine-tuned model with prefix: ``$\texttt{[FINDING]}$ is'' and compare the next-token probability of words meaning `improving', `stable', and `worsening' (\cref{sec:prompting_zero_shot}). 

Without using any labelled data, \Cref{table:temporal_progression_general} shows that the proposed \ac{AR}-based approach already yields performance superior to prior fully-supervised work~\cite{karwande2022chexrelnet} on temporal image classification. With only 10\% of labels, classification fine-tuning provides a further boost, indicating that \temporalbiovil produces a multi-image encoder readily adapted to temporal tasks. Similarly, in a zero-shot report-retrieval setting, the findings show that compared to temporally-agnostic pre-training, \temporalbiovil leveraging prior images improves across all metrics. Consistent with prior work\cite{endo2021retrieval}, the retrieved reports already preserve factuality with high CheXbert scores, more-so than the other metrics which measure fine-grained specifics of phrasing. This demonstrates that the latent space captures the high-level semantics of the clinical features. Fine-grained phrasing however will be substantially improved by AR fine-tuning. 

\begin{table}
    \label{table:report_generation_state_of_the_art}
    \centering
    {
        \notsotiny
        \sisetup{table-format=2.2(2)}
        \caption{
            Report generation results using the same train/test splits from \cite{endo2021retrieval}, measured by lexical (BLEU-2) and factuality (CHEXBERT) metrics. Baseline results were also collected from \cite{endo2021retrieval}. Note the CHEXBERT score covers all 14 observations.
        }
        \resizebox{\linewidth}{!}{
            \begin{tabular}{@{}llSS@{}}
            \toprule
                \textbf{Method} & \textbf{Decoded sections} & \textbf{{BLEU-2}} & {\textbf{CHEXBERT}} \\
                \midrule
                R2gen\cite{chen-emnlp-2020-r2gen} & Findings \& Impression & 21.20 \pm 0.1 & 14.80 \pm 0.3 \\
                IFCC \cite{miura-etal-2021-improving}& Findings & 21.70 \pm 0.10 & 27.00 \pm 0.40 \\
                CXR-RePaiR-Sel \cite{endo2021retrieval} & Impression & 5.00 \pm 0.1 & 27.40 \pm 0.30 \\
                \temporalbiovil & Impression & 15.86 \pm 0.14 & 34.83 \pm0.73 \\
                \temporalbiovil & Findings \& Impression & 21.31 \pm 0.19 & \customBold 35.86 \pm 0.35\\ 
                \bottomrule
            \end{tabular}
        }
    }
    \label{tab:report_generation_state_of_the_art}
\end{table}

\subsection{Achieving \ac{SOTA} performance with \temporalbiovil}\label{sec:sota_performance_biovilt}
\noindent\fbox{\begin{minipage}{23em}
\emph{A wide range of downstream tasks benefit substantially from temporally-aware pre-training.} \end{minipage}} \vspace{.5mm}

Through downstream adaptations and fine-tuning our model, we report \ac{SOTA} performance on report generation and temporal image classification tasks.  For the former, using both prior images \emph{and} reports during fine-tuning substantially improves across metrics (\Cref{table:temporal_decoding_experiments}).
In particular, \ac{TEM} metric results show that temporal context is key for accurately describing change in the generated report while avoiding hallucinations (see \Cref{tab:nlg} for examples). Comparing to published results on a comparable test split and metrics (\cref{tab:report_generation_state_of_the_art}), we conclude that \temporalbiovil with fine-tuning achieves \ac{SOTA} on report generation, producing reports that are lexically on par with prior work but substantially more factually accurate. Note that we do `vanilla' \ac{AR} fine-tuning to focus on the impact of the pre-trained encoders, so application-specific supervision \cite{miura-etal-2021-improving} could be used in conjunction to further boost performance.

In temporal image classification (\cref{table:temporal_progression_general}), \temporalbiovil pre-training outperforms the non-temporal baseline (BioViL) and improves on previously-reported results~\cite{karwande2022chexrelnet} by up to $20$ percentage points (pp). 
Furthermore, baseline methods that rely on image registration (BioViL w/reg), under-perform compared to the proposed approach.
Further analysis reveals that errors tend to be in cases with disagreement between radiologists (\Cref{sec:temporal_image_classification_analysis}). We also note that pre-training is critical for a hybrid CNN-transformer model on this task, likely due to the small labelled dataset.
Lastly, curation of temporal training data is observed to improve the classification results by $.68$ pp aggregated across the findings, see \cref{sec:data_curation_supp} for details.

\subsection{Static tasks benefit from temporal learning}
\label{sec:experiments_static_tasks}

\noindent\fbox{\begin{minipage}{23em}
\emph{\temporalbiovil broadens the range of applicable downstream tasks whilst contributing to performance on static tasks.}
\end{minipage}}\vspace{1.5 mm}

In this section, we demonstrate that performance improvements afforded by \temporalbiovil are not restricted to temporal tasks -- \emph{static} tasks also benefit. 
\Cref{table:rsna_pneumonia_results} reports results on zero- and few-shot pneumonia classification from single images~\cite{shih2019augmenting}, where \temporalbiovil establishes a new \ac{SOTA} compared to prior work~\cite{huang2021gloria,boecking2022making}.

\begin{table}[t]
{\scriptsize
\centering
\sisetup{table-format=1.3}
\caption{\footnotesize Image classification results on RSNA Pneumonia Detection Benchmark \cite{shih2019augmenting} for train and test splits of 70\% -- 30\% respectively.}
\begin{tabular}{@{}lccSSS@{}} 
    \toprule
    \textbf{Method} & \textbf{\% of Labels} & \textbf{Supervision} & \textbf{{Acc.}} & \textbf{{F1}} & \textbf{{AUROC}} \\
    \midrule
    GLoRIA \cite{huang2021gloria}               & \xmark  & Zero-shot   & 0.70  & 0.58  & {-} \\
    \biovil \cite{boecking2022making}           & \xmark  & Zero-shot   & 0.732 & 0.665 & 0.831 \\
    \temporalbiovil                             & \xmark  & Zero-shot   & \bfseries 0.805 & \bfseries 0.706 & \customBold 0.871 \\
    \midrule
    \biovil \cite{boecking2022making}           & 1\%    & Few-shot   & 0.805 & 0.723 & 0.881 \\
    \temporalbiovil                             & 1\%    & Few-shot   & \bfseries 0.814 & \bfseries 0.730 & \customBold 0.890 \\
    \bottomrule
\end{tabular}
\label{table:rsna_pneumonia_results}}
\end{table}

We see a similar trend on the \mscxrbenchmark{} phrase grounding benchmark (\cref{table:ms-cxr-benchmark-results}). This task can be solved with single images, however we show that the inclusion of the prior image (where available) does not impair the performance of \temporalbiovil{}. Feature decomposition effectively preserves localised information from the current image.

\begin{table}[t]
{\footnotesize
\centering
\caption{\footnotesize Results on \mscxrbenchmark{} benchmark \cite{ms-cxr-benchmark} ($5$-runs with different seeds), ``Multi-image'' column indicates the input images used at test time.}
\resizebox{\linewidth}{!}{
\begin{tabular}{@{}lccc@{}}
    \toprule
    \textbf{Method}                       &\textbf{Multi-Image} & \textbf{Avg. CNR} & \textbf{Avg. mIoU}  \\
    \midrule
    \biovil \cite{boecking2022making}     &\xmark     & 1.07 $\pm$ 0.04    & 0.229 $\pm$ 0.005  \\
    + Local loss \cite{boecking2022making,huang2021gloria}   &\xmark    & 1.21 $\pm$ 0.05    & 0.202 $\pm$ 0.010    \\
    \temporalbiovil                       &\xmark     & \bfseries 1.33 $\pm$ 0.04    & \bfseries 0.243 $\pm$ 0.005  \\
    \temporalbiovil                       &\cmark      & \bfseries 1.32 $\pm$ 0.04    & \bfseries 0.240 $\pm$ 0.005  \\
    \bottomrule
\end{tabular}
\label{table:ms-cxr-benchmark-results}
}}
\end{table}

\vspace{-1 mm}
\subsection{Towards better sentence embedding quality}
\noindent\fbox{\begin{minipage}{23em}
\emph{Language models acquire increased temporal sensitivity.}
\end{minipage}}\vspace{1.5 mm}

We hypothesise that text encoders learn temporal semantics through supervision from longitudinal image series. To verify this, RadNLI \cite{miura-etal-2021-improving} and \cxrtbenchmark{} datasets are used in a zero-shot binary classification setting. Cosine similarity of sentence pair embeddings \cite{reimers2019sentence} are treated as class-logits to label each pair either as paraphrase or contradiction. See \cref{sec:sentence_similarity_sup} for further details.

Our text model is benchmarked against SOTA domain-specific BERT models. \Cref{table:ms-cxr-t-sentence-similarity} shows that the proposed framework greatly increases the sensitivity of sentence embeddings to temporal content whilst better capturing the static content (RadNLI). Note that CXR-BERT-Specialised \cite{boecking2022making} is learnt through single-images starting from the same canonical model, illustrating the substantial increase in temporal and static sensitivity due to \temporalbiovil pre-training.

\begin{table}[]
\caption{Results on \cxrtbenchmark{} sentence similarity benchmark.}
{\scriptsize
\centering
\resizebox{\linewidth}{!}{
        \begin{tabular}{@{}lllll@{}}
            \toprule
            & \multicolumn{2}{c}{\textbf{MS-CXR-T} (361 pairs)} & \multicolumn{2}{c}{\textbf{RadNLI} (145 pairs)} \\
            \cmidrule(lr){2-3}\cmidrule(lr){4-5}
            \textbf{Text Model} & \textbf{Accuracy} & \textbf{ROC-AUC} & \textbf{Accuracy} & \textbf{ROC-AUC}  \\
            \midrule
            PubMedBERT~\cite{gu2021domain}        & 60.39  & .542 & 81.38 & .727 \\
            CXR-BERT-G~\cite{boecking2022making}  & 62.60  & .601 & 87.59 & .902 \\
            CXR-BERT-S~\cite{boecking2022making}  & 78.12  & .837 & 89.66 & .932 \\
            \temporalbiovil\                      & \bfseries 87.77 $\pm$ 0.5 & \bfseries .933 $\pm$ .003 & 90.52 $\pm$ 1.0 & \bfseries .947 $\pm$ .003 \\
            \bottomrule
        \end{tabular}
    }
    \label{table:ms-cxr-t-sentence-similarity}
}
\end{table}




\subsection{Ablation experiments}
\label{sec:ablations_and_analyses}
In \Cref{tab:ablation_experiments} we report extensive ablations across the multi-image encoder architecture, pre-training choices, and \ac{AR} fine-tuning for report generation.
\paragraph{Image encoder}

\label{sec:disentangled_features_ablation} \label{sec:bert_inputs_ablation}
\Cref{tab:ablation_experiments} shows that decomposition of static and progression features is essential to ensure good performance on single-image tasks, such as phrase grounding. For temporal representations, on the other hand, positional encodings ($\temporalenc$) are essential to disambiguate the order of scans, i.e., permutation variance across time.
\vspace{1 mm}

\paragraph{Model pre-training}
The corresponding results are shown in the middle section of \Cref{tab:ablation_experiments}. The local contrastive loss proves crucial to ensure meaningful language supervision during pre-training, followed by the image-guided \ac{MLM} objective. Lastly, use of the \Findings{} section results in only minor performance gains as the key findings are already captured in the \Impression{} section.

\newcolumntype{Y}{>{\centering\arraybackslash}X}

\begin{table}
    \scriptsize
    
    \caption{Ablation study on image encoder, pre-training settings, and report generation (one component at a time, and repeated for 4 random seeds). Note that for temporal classification, linear probing is applied to frozen image embeddings. In report generation, the baseline method is fine-tuned with both prior image and report.}
    
    \centering
    \begin{tabularx}{\linewidth}{
        c
        >{\hsize=1.25\hsize\linewidth=\hsize}X
        >{\hsize=.9\hsize\linewidth=\hsize}X
        >{\hsize=.85\hsize\linewidth=\hsize}Y
    } 
        
        \toprule
        & \textbf{Ablation} & \textbf{Avg. CNR (mIoU)} & \textbf{Pl. Effusion Acc.}  \\
        \midrule
        
        \parbox[t]{1mm}{\multirow{3}{*}{\rotatebox[origin=c]{90}{\myuline{Encoder}}}}
        &Baseline                    & 1.33 $\pm$ 0.02 (.248) & 64.8 $\pm$ 0.6 \\
        &\negmark{} Temporal pos. encoding  & 1.32 $\pm$ 0.02 (.242) & 62.9 $\pm$ 1.0 \\
        &\negmark{} Feature decomposition   & 1.11 $\pm$ 0.08 (.203) & 64.0 $\pm$ 0.6 \\
        
        \\ [-1.5ex]
        \hline
        \\ [-1.5ex]
        
        \parbox[t]{1mm}{\multirow{4}{*}{\rotatebox[origin=c]{90}{\myuline{Pre-training}}}}
        &Baseline                     & 1.33 $\pm$ 0.02 (.248) & 64.8 $\pm$ 0.6 \\
        & \negmark{} Use of findings section & 1.32 $\pm$ 0.01 (.246) & 63.8 $\pm$ 0.8 \\
        & \negmark{} MLM loss         & 1.28 $\pm$ 0.02 (.238) & 63.2 $\pm$ 0.7 \\
        & \negmark{} Local contrastive loss & 1.18 $\pm$ 0.02 (.236) & 60.2 $\pm$ 0.6
        \vspace{1 mm}
    \end{tabularx}
    
    \begin{tabularx}{\linewidth}{
        c
        >{\hsize=1.25\hsize\linewidth=\hsize}X
        >{\hsize=.9\hsize\linewidth=\hsize}Y
        >{\hsize=.85\hsize\linewidth=\hsize}Y
    }
        \toprule
         & \textbf{Ablation}    &\textbf{ROUGE} &  \textbf{TEM} \\
        \midrule
        \parbox[t]{1mm}{\multirow{5}{*}{\rotatebox[origin=c]{90}{\myuline{Report gen.}}}}
         & Baseline  &    29.64 $\pm$ 0.08                   &  17.54 $\pm$ 0.11                   \\

         & \negmark{} Prior image &29.35 $\pm$ 0.25& 16.30 $\pm$ 0.40\\
         & \negmark{} Prior report &28.67 $\pm$ 0.12 & 16.00 $\pm$ 0.30\\
         & \negmark{} (Prior image and report) &27.78 $\pm$ 0.09& 13.65 $\pm$ 0.48\\
         & \negmark{} Separation token &26.00 $\pm$ 0.40& 15.50 $\pm$ 1.06\\
        \bottomrule
    \end{tabularx}
    
    \label{tab:ablation_experiments}

\end{table}





\paragraph{Report generation}
The importance of prior image and report is demonstrated by the substantial drop in the ``no prior image and report'' ablation, confirming our hypothesis that temporal context is crucial for improving report quality. While both inputs are crucial for optimal performance, the prior report is more so because it summarises the image and provides a clearer signal. The prior image however cannot be dismissed entirely as it provides granular details which may not always be documented in a report. Finally, we found the separation token is crucial in differentiating between the predicted tokens for the current report and tokens from the prior report. 

\newcommand{\deltapriorimgloss}{\Delta^{\mathrm{prior}}_{\mathrm{img}}}

\subsection{Which tokens require a prior image in MLM?}

\label{sec:token_mlm_analysis}
\begin{figure}[ht]
    \centering
    \includegraphics[width=\linewidth]{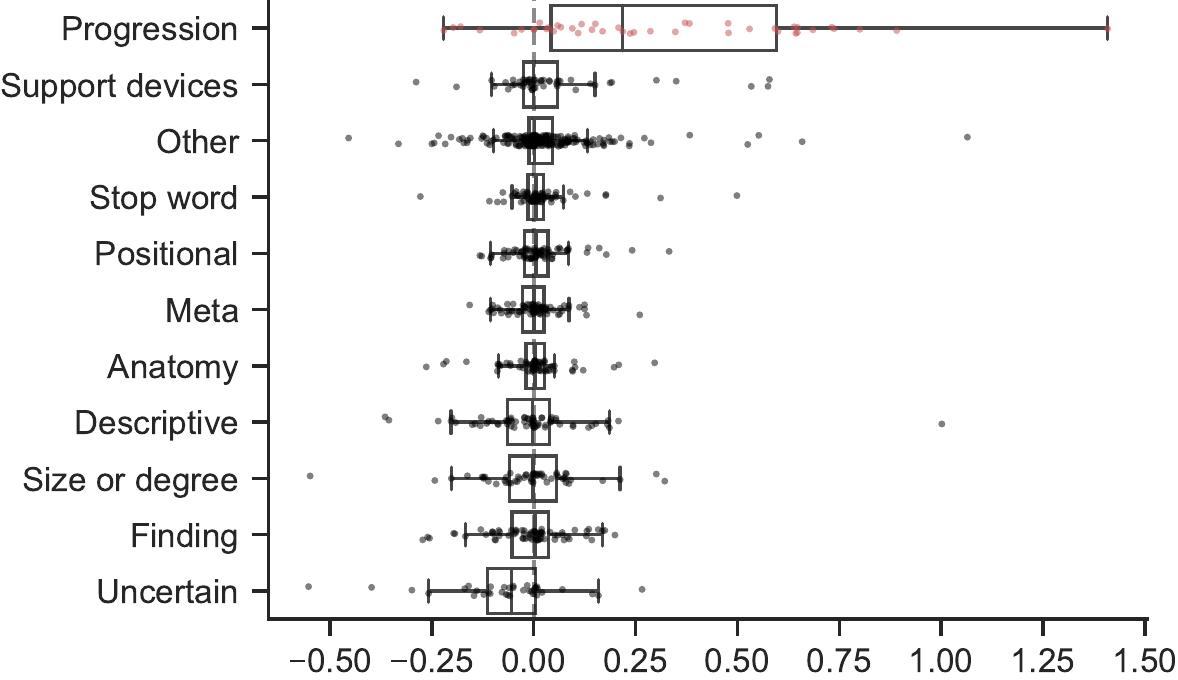}
    {\footnotesize $\deltapriorimgloss(w)$}
    \vspace{-1mm}
    \caption{
        Mean token-level increase in image-guided \ac{MLM} loss when prior image is discarded, grouped by token category.
        The prior image is excluded during inference to measure its impact on masked token predictions.
        \emph{Progression} tokens are significantly better predicted when prior images are incorporated into image embeddings.
        The top five \emph{Progression} tokens are `persist', `improving', `remains', `unchanged', and `residual'. 
    }
    \label{fig:mlm_token-analysis}
\end{figure}

We leverage the \ac{MLM} objective in an inference setting to analyse the influence of prior images in predicting masked tokens. Inspired by the $\Delta$ \emph{image loss} of \cite{bitton2021data}, we define $\deltapriorimgloss$ as the change in loss by conditioning the estimation with a prior image for a given token $w$ as follows:
\begin{equation}
    \deltapriorimgloss(w) = l(w, \currimg, \varnothing) - l(w, \currimg, \previmg)
\end{equation}
where $l(w, \currimg, \previmg)$ is the cross-entropy of predicting the masked token $w$ given visual features (\ac{MLM} loss for a single token), averaged over sentences in which $w$ appears.
$\deltapriorimgloss$ is a measure of how much that token benefits from access to the prior image, as well as an assessment of the contribution of the prior image to the image representation.
In \Cref{fig:mlm_token-analysis} we show the distribution of $\deltapriorimgloss$ as a function of token category
(e.g., \emph{Anatomy}, \emph{Positional}; see \ref{sec:image_guided_mlm} for annotation details).
For \emph{Progression}-type terms in particular, the model heavily relies on the prior image for image-guided \ac{MLM}. We further observe that this effect is specific to temporal tokens; as expected, those from other semantic categories do not consistently rely on the prior image.

%% file: content/figure_tex/fig_attention.tex
\newcommand{\cxrwidth}{0.11\textwidth}

\begin{figure}
     \centering
     \begin{subfigure}[b]{\cxrwidth}
         \centering
         \caption*{Prior image}
         \includegraphics[width=\textwidth]{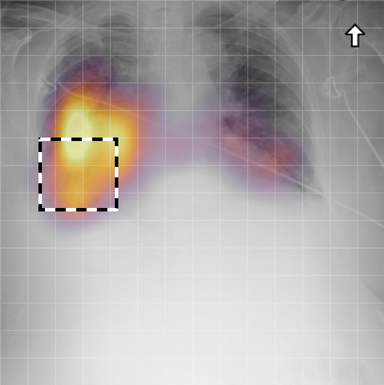}
     \end{subfigure}
     \begin{subfigure}[b]{\cxrwidth}
         \centering
         \caption*{Current image}
         \includegraphics[width=\textwidth]{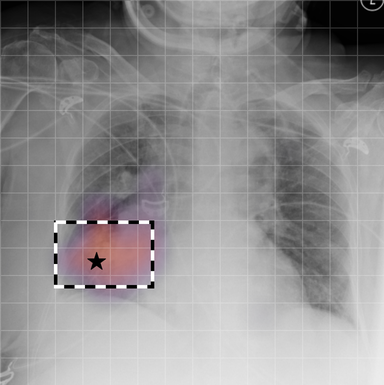}
     \end{subfigure}
     \hfill
     \begin{subfigure}[b]{\cxrwidth}
         \centering
         \caption*{Prior image}
         \includegraphics[width=\textwidth]{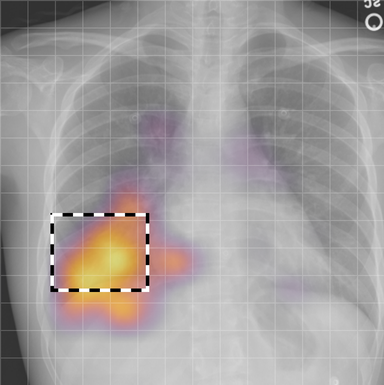}
     \end{subfigure}
     \begin{subfigure}[b]{\cxrwidth}
         \centering
         \caption*{Current image}
         \includegraphics[width=\textwidth]{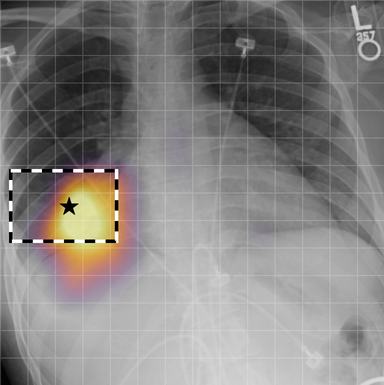}
     \end{subfigure}
     \caption{
        Attention rollout maps~\cite{abnar-zuidema-2020-quantifying} from the reference patch (marked with $\filledstar$) to the current and prior images.
        The bounding boxes, annotated by a radiologist, show the extent of consolidation.
        Note that the reference patch attends to its anatomical neighbourhood in the prior image despite the misalignment between prior and current images.
        The grid ($14 \times 14$) represents the patch tokens processed in the transformer encoder blocks.
    }
    \label{fig:self_attention_pose_variations}
\end{figure}

%% file: content/conclusion.tex
\section{Conclusion}
In this paper, we introduced \temporalbiovil, a vision--language pre-training framework enabling alignment between text and multiple images. 
\temporalbiovil makes use of a novel multi-image encoder and explicitly decomposes static--temporal features to augment the current image representation with information from prior images. This enables the grounding of temporal references in the text.
To our knowledge, this is the first method capable of leveraging the temporal content commonly present in biomedical text. It addresses an important limitation in existing \ac{VLP} approaches, which simply discard such context.
Also, incorporating such multi-modal temporal content provides strong learning signals to the model, resulting in richer representations and improved downstream performance.

We demonstrate the value of this paradigm through extensive experiments: \temporalbiovil excels on both static and temporal tasks, establishing new \ac{SOTA} on report generation, temporal image classification, few/zero-shot pneumonia detection, and phrase grounding. Furthermore, we release a new multi-modal benchmark (\cxrtbenchmark) to measure the quality of image and text representations in terms of temporal semantics, enabling more diverse evaluation of biomedical VLP models. The corresponding model weights\footnote{Models can be found at: \modelurl} and code\footnote{Code can be found at: \sourcecodeurl} are publicly available.

Further exploration and evaluation are required on diverse datasets to characterise what kinds of tasks would benefit from a temporal modelling approach, and specifically from the proposed methodology.

\paragraph{Acknowledgements:} We would like to thank Hannah Richardson, Hoifung Poon, Melanie Bernhardt, Melissa Bristow and Naoto Usuyama for their valuable feedback.

%% file: content/supplementary_material.tex
\beginsupplement
\appendix
\input{content/figure_tex/commands_supp}

\section{Additional Results and Analyses}

\subsection{Qualitative analysis of generated reports}
\label{sec:nlg_examples}
\input{content/figure_tex/table_report_examples}

\Cref{tab:nlg} shows example reports generated with \temporalbiovil and \biovil models, which are compared to the reference radiologist's reports. In comparison with \biovil which only models the current image, \temporalbiovil shows the benefit from incorporating prior study information and is able to provide factually more accurate reports especially in terms of describing temporal progression of the findings. This is showcased in the first two examples in the table: In the first row, \temporalbiovil is able to comment on not only the presence of the pleural effusion but also its improvement  while \biovil fails to mention the change. In the second example, \temporalbiovil is able to correctly identify that there is no relevant change by comparing with the previous study, while \biovil wrongly hallucinates the tube in the current image as a new placement. \temporalbiovil can also avoid hallucination of the temporal information when there is no prior study. For instance, in the third example, \temporalbiovil correctly acknowledges that there is no prior image and generates the report based on information from the single current image, while \biovil hallucinates a non-exisistent prior study and wrongly generates temporal descriptions in the report. 

\subsection{Further analysis on temporal classification}
\label{sec:temporal_image_classification_analysis}
A subset of the \cxrtbenchmark{} benchmark dataset is re-annotated by an expert radiologist by blinding them to the existing ground-truth labels and displaying only pairs of images obtained from each subject. With the new set of labels, the analysis focuses on measuring the correlation between inter-rater agreement and image model's prediction errors. \Cref{fig:cxrtloss} shows the dependency between the two where the x-axis corresponds to the cross entropy loss between the \cxrtbenchmark{} benchmark labels and model predictions. We observe lower model performance on cases with smaller inter-rater reliability for the three classes in the dataset, indicating that the model's prediction errors occur more often for the cases where experts may disagree with each other.

\input{content/figure_tex/fig_cxr-t_loss_label_quality}

\subsection{Self-attention visualisation}
\label{sec:attention_viz}
In \Cref{fig:self_attention_qualitative_examples}, we show examples  of self-attention rollout~\cite{abnar-zuidema-2020-quantifying} maps for pleural effusion and consolidation, including radiologist-annotated bounding boxes surrounding the corresponding pathology in each prior and current image.

To model the attention flow through the transformer encoder block, we first average each attention weight matrix across all heads, subsequently we multiply the matrices between every two layers. For every block we add the identity matrix in order to model the residual connections. Last, we only keep the top 10 $\%$ of attention weights per block to reduce noise in the final rollout map. In contrast to \cite{dosovitskiy2020image}, we do not visualize the rollout map with respect to a $\clstoken$ token. Instead, we choose a reference image patch from the center of the radiologist-annotated bounding boxes, marked with $\filledstar$ in \Cref{fig:self_attention_qualitative_examples}. 

We find that the rollout maps in \Cref{fig:self_attention_qualitative_examples} are in good agreement with radiologist-annotated bounding boxes, i.e., the reference patch attends to other patches within the bounding boxes in the prior and current image. In addition, we find that \temporalbiovil{} is robust to pose variations, e.g., in \Cref{fig:self_attention_qualitative_examples} (a) we show that despite the vertical shift between prior and current image, the reference patch attends to the correct image patches in the prior image.

To further assess the robustness of \temporalbiovil{} against pose variations between prior and current images, we performed multiple rotations to the prior image within a pair and computed rollout maps from the same reference patch in the current image.
\Cref{fig:self_attention_pose_variations_supp} shows that \temporalbiovil{} consistently attends to the corresponding anatomical region independently of the spatial transformation applied, demonstrating that registration is not needed.

\input{content/figure_tex/fig_attention_supp}

\subsection{Data curation of imaging datasets}
\label{sec:data_curation_supp}

Large datasets often contain instances that are mislabelled or out of distribution \cite{jia2021scaling}.
We used \temporalbiovil{} to perform pairwise ranking of instances in MIMIC-CXR (\Cref{sec:adaptations_to_downstream}, $\delta = 0.2$) and selected representative examples found in the dataset. Our method is able to select the most appropriate image for a range of different image-acquisition or image-processing issues (\Cref{fig:data_curation_supp}).

We found that many lateral acquisitions in the dataset were unexpectedly labelled as frontal (\Cref{fig:auto_qc_view}).
Some images contained only noise (\Cref{fig:auto_qc_noise}), non-human samples (\Cref{fig:auto_qc_object_1,fig:auto_qc_object_2}) or incorrect anatomy (\Cref{fig:auto_qc_leg}).
Often, acquisitions with an incomplete \ac{FOV} (i.e., the lungs are not completely visible) were repeated (\Cref{fig:auto_qc_fov}). Lastly, post-processed images were detected by the algorithm such as contrast-enhanced scans (\Cref{fig:auto_qc_clahe}) that are not often used for diagnostic purposes in clinical practice.

\input{content/figure_tex/fig_data_curation}

\subsection{Phrase-grounding on external data}
We have additionally conducted a robustness analysis on an out-of-distribution dataset. For this purpose, a small set of expert labels (N=$137$ bounding-box--caption pairs) were collected on Open-Indiana CXR dataset \cite{demner2016preparing} for phrase grounding on the same set of abnormalities as \mscxrbenchmark{} benchmark \cite{ms-cxr-benchmark}. The dataset differs in terms of text token distribution, demographics, and disease prevalence. The experiment was performed with the same methods and setup described in \Cref{sec:experiments_static_tasks}. The results show that the performance gains due to temporal pre-training is observed to be consistent on external datasets.

\begin{table}[h!]{
\scriptsize
\centering
\caption{Multi-modal phrase-grounding results obtained on a subset of Open-Indiana CXR dataset \cite{demner2016preparing} image-text pairs. ``Multi-image'' column indicates the input images used at test time. The results are reported in terms of micro-averages owing to the limited number of samples in some classes.}
\begin{tabular}{@{}lcccc@{}}
    \toprule
    \textbf{Method}    &\textbf{Pre-Train} &\textbf{Multi-Image} & \textbf{Avg. CNR} & \textbf{Avg. mIoU}  \\
    \midrule
    \biovil [9]       &   Static           &\xmark    & 1.19 $\pm$ 0.04    & 0.259 $\pm$ 0.003  \\
    \temporalbiovil   &   Temporal        &\xmark    & \bfseries 1.53 $\pm$ 0.05    & \bfseries 0.289 $\pm$ 0.006  \\
    \bottomrule
\end{tabular}\par}
\vspace{-5mm}
\end{table}

\section{Temporal aspects of the MIMIC-CXR v.2 dataset}
\label{sec:mimic_appendix_longitudinal}

\begin{figure}
    \centering
    \includegraphics[width=0.45\textwidth]{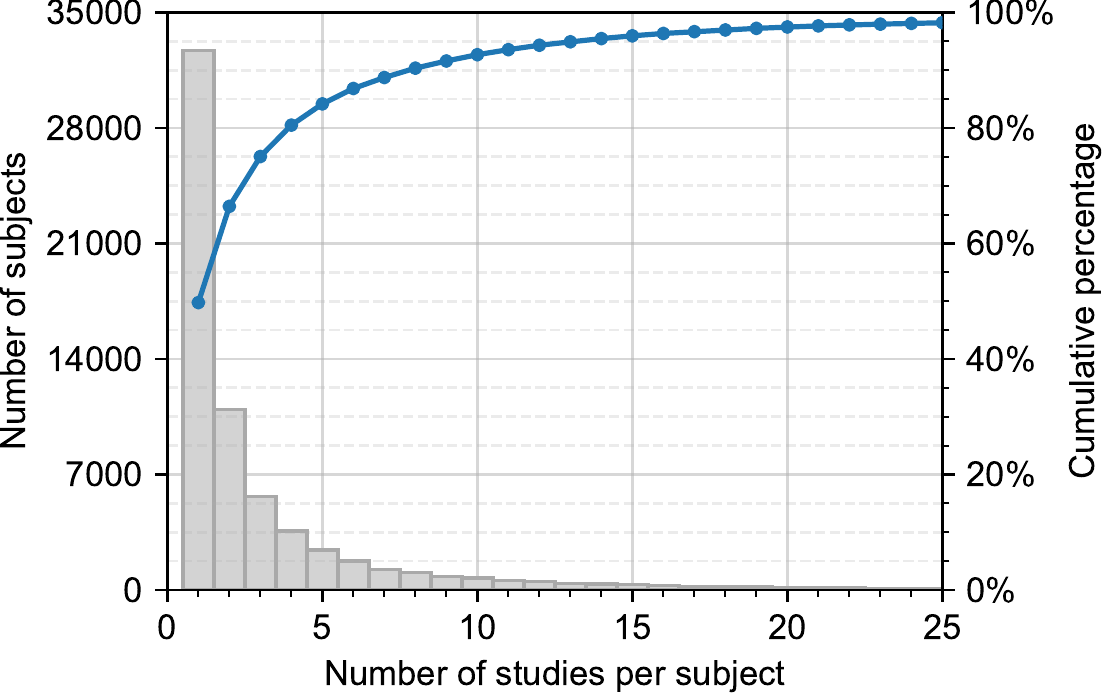}
    \caption{
        Number of studies per subject in the MIMIC-CXR dataset.
        A study, sometimes referred to as an `exam' or `procedure', refers to ``one or more images taken on a single visit to a medical facility'' (adapted from \url{https://ncithesaurus.nci.nih.gov/}).
        Note that 67\,\% of subjects have at least two studies that happened at different times.
    }
    \label{fig:histogram_longitudinal_studies}
\end{figure}

Subjects in the MIMIC-CXR dataset often have multiple associated studies that happened at different times.
A study, sometimes referred to as an `exam' or `procedure', refers to ``one or more images taken on a single visit to a medical facility''%
\footnote{Adapted from \url{https://ncithesaurus.nci.nih.gov/}}.
To assess pathology progression, radiologists compare images (also referred to as `scans' or `series') from different studies.
In the MIMIC-CXR dataset, each study (with one or more images) is accompanied by the report written by the radiologist. \Cref{fig:histogram_longitudinal_studies} represents the distribution of studies per subject within MIMIC-CXR and the corresponding cumulative distribution function, showing that 67\,\% of the subjects have at least two different associated studies (and therefore at least two images acquired at different stages of the disease).

Another way to quantify temporal information in MIMIC-CXR is through the progression labels provided by the Chest ImaGenome dataset \cite{chest-imagenome}. These progression labels are extracted from the reports and thus identify the cases when the radiologist explicitly describes changes. We found that in MIMIC, around 40\,\% of the reports are associated with a progression label from any of the available findings defined by ImaGenome.

\section{\cxrtbenchmark{} benchmark}
\label{sec:cxrt-benchmark}
\subsection{Temporal image classification}

The \cxrtbenchmark\ temporal image classification contains progression labels for five findings (Consolidation, Edema, Pleural Effusion, Pneumonia and Pneumothorax) across three progression classes (\texttt{Improving}, \texttt{Stable}, and \texttt{Worsening}). This benchmark builds on the publicly available Chest ImaGenome gold and Chest ImaGenome silver datasets \cite{chest-imagenome} which provide progression labels automatically derived from radiology reports. We collected a set of studies that are part of the ImaGenome silver dataset, excluding any studies that had been previously verified as part of the ImaGenome gold dataset. Additionally, we excluded studies where there are multiple progression labels for a single pathology (e.g. left pleural effusion has increased, right pleural effusion remains stable). We conducted a review process of the selected candidates, asking a board certified radiologist to either accept or reject the label. To inform their review of the labels, the radiologist was given access to the radiology report for the current image, and the sentence from which the auto generated label had been extracted.

After collecting our curated labels and labels from the ImaGenome gold dataset, we matched the report-based labels to specific image pairs, performing a second data curation step to create the image dataset. To ensure the diagnostic quality of all images in the dataset, if a study had multiple frontal scans we performed a quality control step asking a radiologist to select the best image for each study. \cref{fig:cxrtimgs} shows examples from the benchmark across different pathologies and progression labels.

The class distribution for the image classification task in \cxrtbenchmark\  is shown in \cref{table:progression_benchmark_details}. As seen in the table, the class distribution of the dataset skews towards the  \texttt{stable} and  \texttt{worsening} classes. This could be explained as patients are more likely to get a chest X-ray scan when their condition is stable or deteriorating as opposed to when there is an improvement in patient condition.

\begin{table}[t!]
    \centering
    \caption{
        \cxrtbenchmark\ temporal image classification benchmark: Showing the distribution of multi-image studies across different clinical findings, distribution of classes \{\texttt{Improving}, \texttt{Stable}, \texttt{Worsening}\} per finding, and number of subjects.
    }
    \resizebox{\linewidth}{!}{
    \begin{tabular}{@{}lc r@{ / }r@{ / }r c@{}}
        \toprule
        \bfseries Findings & \bfseries \# of annotation pairs & \multicolumn{3}{c}{\bfseries Class distribution} & \bfseries \# of subjects\\
        \midrule
        Consolidation    & 201 & 14\% & 42\% & 44\%  & 187 \\ 
        Edema            & 266 & 31\% & 26\% & 43\%  & 241 \\ 
        Pleural effusion & 411 & 19\% & 49\% & 32\%  & 370\\ 
        Pneumonia        & 237 & 8\% & 25\% & 67\% & 218 \\ 
        Pneumothorax     & 211 & 15\% & 55\% & 30\% & 148 \\ 
        \midrule
        Total            & 1326 & 18\% & 40\% & 42\% & 800\\
        \bottomrule
    \end{tabular}}{}
    \label{table:progression_benchmark_details}
\end{table}
\begin{table}[t]
    \centering
    \caption{\cxrtbenchmark\ temporal sentence similarity benchmark: Number of paraphrase and contradiction examples in the full dataset and across the \texttt{RadGraph} and \texttt{Swaps} subsets.}
    \resizebox{0.9\linewidth}{!}{
    \begin{tabular}{@{}lc c c c@{}}
        \toprule
        \bfseries Subset & \bfseries \# of paraphrase pairs & \bfseries \# of contradiction pairs & \bfseries Total \\
        \midrule
        Radgraph & 42 & 75 & 117 \\
        Swaps & 99 & 145 & 244 \\
        \midrule
        Total & 141 & 220 & 361 \\
        \bottomrule
    \end{tabular}}{}
    \label{tab:sentence_similarity_benchmark_details}
\end{table}

\subsection{Temporal sentence similarity}
\label{sec:sentence_similarity_supp}
 In this section, we describe the process of creating the \cxrtbenchmark\ temporal sentence similarity benchmark, which consists of pairs of paraphrase or contradiction sentences in terms of disease progression. We create this dataset using two different methods, \texttt{RadGraph} where paraphrase and contradiction sentence pairs are discovered by analysing graph representations of sentences and \texttt{Swaps} where paraphrases and contradictions are created by swapping out temporal keywords in the sentence. 

\begin{table*}[t!]
    \centering
    \caption{Examples of paraphrase and contradiction sentence pairs from the \cxrtbenchmark\ temporal sentence similarity benchmark. The examples are selected from the \texttt{RadGraph} and \texttt{Swaps} subsets (see \Cref{sec:sentence_similarity_supp}). }
    \resizebox{\linewidth}{!}{
    \footnotesize		
    \begin{tabular}
    {@{} p{0.02\linewidth} p{0.10\linewidth} p{0.44\linewidth} p{0.44\linewidth} @{}}
        \toprule
         & \bfseries Label & \bfseries Sentence 1 & \bfseries Sentence 2 \\
         \midrule
        \parbox[t]{3mm}{\multirow{2}*{\rotatebox[origin=c]{90}{\underline{\footnotesize Swaps}}}}
         & \multirow{1}*{Paraphrase} &``Unchanged small-to-moderate right pleural effusion.'' & ``Stable small-to-moderate right pleural effusion.'' \\[2mm]
        
         & \multirow{1}*{Contradiction}  &``Interval worsening of the right-sided pneumothorax.'' & ``Interval resolution of the right-sided pneumothorax.''\\

         \midrule
        \parbox[t]{3mm}{\multirow{2}*{\rotatebox[origin=c]{90}{\underline{\footnotesize RadGraph}}}} & \multirow{1}*{Paraphrase} &``There has also been a slight increase in left basal consolidation.'' & ``There is slight interval progression of left basal consolidation.''\\ [2 mm]
         
         & \multirow{1}*{Contradiction} &``Right mid and lower lung consolidations are unchanged.'' & ``There has been worsening of the consolidation involving the right mid and lower lung fields.''\\
         
        \bottomrule
    \end{tabular}
    }{}
    \label{tab:sentence_similarity_benchmark_examples}
\end{table*}

 To create this dataset, we first collected a set of sentences from the MIMIC dataset, using the Stanza constituency parser \cite{zhang2021biomedical} to extract individual sentences from reports. Using the CheXbert labeller \cite{smit2020chexbert}, we filtered this set to sentences that described one of seven pathologies - Atelectasis, Consolidation, Edema, Lung Opacity, Pleural Effusion, Pneumonia or Pneumothorax. We then filtered to sentences which contained at least one mention of a temporal keyword. Using this sentence pool, paraphrase and contradiction pairs were constructed in two ways.
(I) We paired sentences from the sentence pool by matching on RadGraph \cite{jain2021radgraph_supp} entities, relaxing the matching constraint only for temporal keywords and possible mentions of pathologies.
(II) We swapped out temporal keywords in a sentence to create sentence pairs, choosing swap candidates from the top 5 masked token predictions from CXR-BERT-Specialized \cite{boecking2022making} provided they were temporal keywords.
 After creating candidate sentence pairs, we manually filtered out sentence pairs with ambiguous differences in terms of disease progression. A board certified radiologist then annotated each sentence pair as either paraphrase or contradiction. Sentences were filtered out in the annotation process if (I) they were not clear paraphrases or contradictions (II) the sentences differed in meaning and this difference was not related to any temporal information (III) they were not grammatically correct. The distribution of sentence pairs across the paraphrase and contradiction classes are described in \Cref{tab:sentence_similarity_benchmark_details}, see \Cref{tab:sentence_similarity_benchmark_examples} for examples from the benchmark.

\section{Temporal entity matching}
\label{sec:tem_metric}
To quantify how well the generated report describes progression-related information, we propose a new metric, namely temporal entity matching (\ac{TEM}) score.
\subsection{Metric Formulation}

We first extract entities (tagged as ``observation'' or ``observation\_modifier'') from the text by running the named entity recognition model in the Stanza library \cite{zhang2021biomedical}. Within the extracted entities, we manually curated a list of temporal entities that indicate progression (\Cref{sec:tem_keywords}). The list is reviewed by an expert radiologist. Given extracted temporal entities $E$ in $N$ pairs of reference and generated reports, we calculate global precision ($p_E$) and global recall ($r_E$), which are later used to compute the \ac{TEM} score. It is defined as the harmonic mean of precision and recall (also known as the F1 score). 
\begin{align}
    p_E &= \frac{\sum_{i=1}^N |E_{gen}^{i} \cap E_{ref}^{i}|}{\sum_{i=1}^N |E_{gen}^{i}|} \\
    r_E &= \frac{\sum_{i=1}^N |E_{gen}^{i} \cap E_{ref}^{i}|}{\sum_{i=1}^N |E_{ref}^{i}|}
\end{align}

\subsection{List of temporal keywords}
\label{sec:tem_keywords}
The list of temporal keywords used to compute the TEM score are as follows:
\{bigger,  change,  cleared,  constant,  decrease,  decreased,  decreasing,  elevated,  elevation,  enlarged,  enlargement,  enlarging,  expanded,  greater,  growing,  improved,  improvement,  improving,  increase,  increased,  increasing,  larger,  new,  persistence,  persistent,  persisting,  progression,  progressive,  reduced,  removal,  resolution,  resolved,  resolving,  smaller,  stability,  stable,  stably,  unchanged,  unfolded,  worse,  worsen,  worsened,  worsening, unaltered\}.

\section{Architecture and implementation details}
\label{sec:implementation_details}

\subsection{Hyper-parameters}
\label{sec:hyperparameters}
The models are trained in a distributed setting across 8 GPU cards. For pre-training, we use a batch size of 240 (30 * 8 GPUs) and the AdamW optimizer \cite{loshchilov2018decoupled_supp}. We use a linear learning rate scheduler with a warm-up proportion of 0.03 and base learning rate of $2 \times 10^{-5}$. We train for a maximum of 50 epochs and use validation set loss for checkpoint selection. The overall loss is a sum of components with weighting factors: global contrastive (1.0), local contrastive (0.5), and image-guided \ac{MLM} (1.0) respectively, see \cref{sec:image_features} for further details on their formulation.

Following \cite{boecking2022making} we use sentence permutation as text-based data augmentation. Similarly, spelling errors in the reports are corrected prior to tokenisation of the text data\footnote{\url{https://github.com/farrell236/mimic-cxr/blob/master/txt/section_parser.py}}. For image augmentations, note that we apply the same augmentation to current and prior images to prevent severe mis-alignment. We resize the shorter edge to 512 and centre-crop to (448, 448). We apply random affine transformations (rotation up to $30^\circ$ and shear up to $15^\circ$) and colour jitter (brightness and contrast).

\subsection{Training infrastructure}
We train with distributed data processing (DDP) on eight NVIDIA Tesla V100s with 32GB of memory each. To handle inconsistently-present prior images with DDP, we define a custom batch sampler. This sampler is a mixture of two samplers, in proportion to their dataset coverage: a sampler which produces batches with \emph{only} multi-image examples -- $(\currimg, \previmg, \currreport) \in \multiimageset$ and one with only single-image examples -- $(\currimg, \varnothing, \currreport) \in \singleimageset$. Each GPU then processes a batch which is entirely single or multi-image, avoiding branching logic within the forward pass and enabling an efficient single pass through the CNN to process all input images (current or prior) by concatenating them along the batch dimension.

We confirmed that although the custom sampler theoretically impacts the order in which the dataset is traversed, it has a negligible effect on training metrics relative to fully random sampling. Since we train on eight GPUs and collect negatives across all GPUs during contrastive training, each update involves on average a representative mixture of both single-image and multi-image samples.

Finally, following \cite{boecking2022making} we use the DICOM images from MIMIC-CXR to avoid JPEG compression artefacts.

\section{Adaptation and experimentation details}
\label{sec:experiment_details}
\subsection{Fine-tuning \temporalbiovil for report generation}
\label{sec:report_generation_details}
During fine-tuning of \temporalbiovil for report generation, we minimise the cross entropy loss to maximise the log likelihood of the report in an autoregressive manner given the input images. The model is initialised from the pretrained weights of the image encoder and the text encoder. Similar to the cross-modal masked language modelling task, we additionally train a linear projection layer to map the projected patch embeddings to the same hidden dimension of the text encoder, and we train cross-attention layers in each transformer block. The difference from the masked language modelling task is that we change the bidirectional self-attention to unidirectional causal attention that can only access the past tokens. If trained with prior report, we pass the prior report as prefix to condition the generation of the current report (the current and prior report are separated by $\septoken$), and we only back-propagate the gradients from the loss on the tokens in the current report. 

For all experiments, we train the model for 100 epochs and we chose the best checkpoint according to metrics on the validation set. We performed grid search for learning rate in $[10^{-5}, 2 \times 10 ^{-5}, 5 \times 10 ^{-5}]$ and found $2 \times 10 ^{-5}$ to be optimal.
We ran each experiment with 3 random seeds and report mean and standard deviation. 

In addition to the metrics we reported in the main text, we also evaluate the generated reports by \ac{NEM}. This metric was defined in \cite{miura-etal-2021-improving} to measure the accuracy of reporting clinically relevant entities in the generated reports (Similar to how \ac{TEM} is computed to measure the match of temporal entities in our study). Following \cite{miura-etal-2021-improving}, we extract entities (tagged as ``observation'' or ``observation\_ modifier'') from the text by running the named entity recognition model in the Stanza library \cite{zhang2021biomedical}. The results are presented in \cref{table:temporal_decoding_experiments_nem}.

\begin{table}[t]
    \centering
    \caption{Results for report generation task: Predictions are evaluated on NEM. The approaches are grouped into two broad categories: NN (Nearest Neighbour) and AR (Auto-Regressive). \temporalbiovil\ pre-training consistently yields superior decoding performance. Further, the use of prior image and report consistently yield performance gains demonstrating the importance of such domain priors.}
    \setlength{\tabcolsep}{4pt}
    \resizebox{\linewidth}{!}{
    
    \sisetup{table-format=2.2(1)}
    \begin{tabular}{@{}l@{\hskip 0.2cm}llc S@{}}

        \toprule
        & \bfseries Method                   & \bfseries Pre-training & \bfseries Prior Img/Report  &{\bfseries NEM} \\
        \midrule
        \parbox[t]{3mm}{\multirow{3}{*}{\rotatebox[origin=c]{90}{\underline{NN}}}}
        &CXR-RePaiR-2 \cite{endo2021retrieval}      &   \biovil\            &   \xmark\ /  \xmark  &
        13.36   \\
        
        &Baseline (NN) \cite{boecking2022making}    &   \biovil\            &   \xmark\ /  \xmark  & 
        16.25   \\
        
        &Proposed (NN)                              &   \temporalbiovil\ & \cmark / \xmark
        & 17.55   \\
    
        \midrule
        \parbox[t]{3mm}{\multirow{3}{*}{\rotatebox[origin=c]{90}{\underline{AR}}}}
        &Baseline (AR) \cite{boecking2022making}    & \biovil\        & \xmark\ / \xmark  &   24.27 \pm 0.22 \\
        &Proposed                                   & \temporalbiovil & \cmark / \xmark &   25.50 \pm 0.04   \\
        &Proposed                                   & \temporalbiovil & \cmark / \cmark   &   \customBold 26.95\pm0.17                                       \\
        \bottomrule
    \end{tabular}}
    \label{table:temporal_decoding_experiments_nem}
\end{table}

\subsection{Nearest-neighbour-based report retrieval}
\label{sec:nearest_neighbour_details}
The joint latent space learnt by \temporalbiovil can also be used to directly perform report retrieval without requiring task-specific model fine-tuning. Given the test image, we retrieve its semantically closest report from the training set in the joint latent space. Specifically, we encode each test image with the image model in \temporalbiovil and collect its projected image embeddings, and similarly we encode all the reports in the training data with their projected text embeddings. For each test study, we compute cosine similarity between the test image embedding and all the text embeddings from the training set in the joint latent space, and we retrieve the closest text embedding and use its corresponding report as the prediction. To evaluate the retrieval performance, we use the same decoding metrics on the retrieved reports and report results in the top section of \Cref{table:temporal_decoding_experiments}. In a separate set of experiments, we also tried performing nearest neighbour search only within the image embedding space by retrieving the report associated with the closet image embedding, but this yielded sub-optimal performance compared with using the joint latent space. 

\subsection{Fine-tuning for temporal image classification}
\label{sec:tic_details}

In this section, we describe the training dataset and fine-tuning procedure for the fully supervised and few-shot settings of the temporal image classification task. For this task, we finetune \temporalbiovil on a subset of the Chest ImaGenome silver dataset \cite{chest-imagenome} to predict progression labels for 5 different pathologies. To create our training dataset, we filter out image pairs from this dataset where there are multiple directions of progression of a single pathology in the image-pair.  We additionally perform an automatic data curation step to choose higher quality image pairs when possible, as described in \ref{sec:adaptations_to_downstream}. \Cref{table:imagenome_silver} shows the number of training samples and label distribution for the training dataset.

\begin{table}[th!]
    \centering
    \caption{
        Statistics of the training dataset used for downstream fine-tuning on temporal image classification.
    }
    \resizebox{\linewidth}{!}{
    \begin{tabular}{@{}lc r@{ / }r@{ / }r c@{}}
        \toprule
        \bfseries Findings & \bfseries \# labelled pairs & \multicolumn{3}{c}{\bfseries Class distribution} & \bfseries \# of subjects\\
        \midrule
        Consolidation    & 7012 & 15\%  & 42\% & 43\%  & 3308 \\ 
        Edema            & 14170 & 28\%  & 33\% & 39\%  & 4813 \\ 
        Pleural effusion & 26320 & 16\%  & 53\% & 31\%  & 6838\\ 
        Pneumonia        & 8471 & 12\%   & 29\% & 59\% & 4197 \\ 
        Pneumothorax     & 3795 & 21\%  & 57\% & 22\% & 1161 \\ 
        \bottomrule
    \end{tabular}}{}
    \label{table:imagenome_silver}
\end{table}

For the fully supervised setting, we add a multilayer classification head to the \temporalbiovil image encoder and fine-tune the model independently for each pathology. We use weighted cross entropy loss with a batch size of 128 and the AdamW optimizer \cite{loshchilov2018decoupled_supp}. During parameter optimisation, positional encodings and missing-image embeddings are exempt from weight decay penalty as in \cite{wu2021rethinking_supp}. We train for 30 epochs, with a linear learning rate schedule, a warmup proportion of $0.03$ and a base learning rate of $1 \times 10^{-5}$. For data augmentation, we first resize the shorter edge of the image to 512 and centre crop to (448, 448). We apply random horizontal flips, random cropping, random affine transformations (rotation up to $30^\circ$, shear up to $15^\circ$), colour transforms (brightness and contrast) and Gaussian noise. 

For the few-shot setting we tune only a single-layer linear head on the \temporalbiovil image encoder and freeze the rest of the encoder. We initialise the weight matrix of the linear head with values from encoded text prompts \cite{boecking2022making} for each of the three progression classes, and the bias matrix is initialised with zeros. To train, we again use weighted cross entropy loss, with a batch size of 32 and the AdamW optimizer. We use a learning rate of $1 \times 10^{-3}$ and train for 40 epochs. 
For data augmentation, we resize the shorter edge of the image to 448 and center crop to (448, 488).  We apply random horizontal flips, random affine transformations (rotation up to $45^\circ$ and shear up to $25^\circ$), colour transforms (brightness and contrast). As in the pre-training step, we always synchronise image data augmentations to apply the identical transforms to the current and prior images.

\subsection{Auto-regressive prompting for zero-shot temporal image classification}
\label{sec:prompting_zero_shot}
Following the GPT-3 style language prompting \cite{brown2020language}, we prompt the fine-tuned AR language decoding model with the template:
``$\texttt{[FINDING]}$ is'' and infer the next token to perform temporal classification for each of the five findings. The mapping from the predicted next token to the three progression classes is characterised by a short list of tokens provided in \Cref{tab: language_decoding_verbalizer}. After computing the posterior for each token in the list, the obtained values are normalised across the three classes, and the class with the highest score is selected as the prediction. The corresponding results are reported in \Cref{table:temporal_progression_general}.

\begin{table}
\caption{Prompting the AR language decoding model for zero-shot image classification. The list above shows the mapping from decoded tokens to progression classes. }
    \centering
    {\footnotesize
    \begin{tabular}{p{0.1\textwidth} p{0.27\textwidth}   }
    \toprule
    \bfseries Target class & \bfseries Tokens \\
    \midrule
    Improving & better, cleared, decreased, decreasing, improved, improving, reduced, resolved, resolving, smaller\\ 
    \midrule
    Stable & constant, stable, unchanged\\
    \midrule
    Worsening &  bigger, developing, enlarged, enlarging, greater, growing, increased, increasing, larger, new, progressing, progressive, worse, worsened, worsening \\
    
    \bottomrule
    
\end{tabular}}
    \label{tab: language_decoding_verbalizer}
\end{table}

\begin{table*}[t!]
    \centering
    \begin{tabular}{l l l}
    \toprule
    \textbf{Category} & \textbf{Description}  & \textbf{Examples} \\
    \midrule
    Progression & Pertaining to change or progression & \emph{bigger, cleared, new} \\
    Support devices & Tubes, lines and implants & \emph{nasogastric, pacemaker, cannula}\\
    `Other' & No clear category & \emph{can, relevant, overall}\\
    Stop word & `Insignificant' words & \emph{the, no, of}\\
    Positional & Localisation (not anatomical) & \emph{right, lower, bilateral}\\
    Meta & Pertaining to the report itself or practice of radiology & \emph{evidence, radiograph, study} \\
    Anatomy & Anatomical locations & \emph{pulmonary, chest, mediastinal}\\
    Descriptive & Qualitative appearance of a finding & \emph{layering, focal, patchy}\\
    Size or degree & Quantifying extent or severity & \emph{extensive, moderate, severe}\\
    Finding & Radiographic finding or pathology & \emph{edema, penumonia, pneumothorax}\\
    Uncertain & Expression of certainty or doubt & \emph{may, possible, concerning}\\
    \bottomrule
    \end{tabular}
    \caption{Semantic categories used in \Cref{fig:mlm_token-analysis}.}
    \label{tab:semantic_categories}
\end{table*}

\subsection{Further analysis of image-guided \ac{MLM}}
\label{sec:image_guided_mlm}
In \Cref{sec:token_mlm_analysis} we used a simplified notation for the computation of $\deltapriorimgloss(m)$ for ease of exposition -- here we provide further detail. Recall that $\phrase=(w_1,\dots,w_\numtexttokens)$ is a sequence of tokens and $\phrase_{\backslash m}$ is that sequence with token $m$ masked. Let $p_\theta(\phrase_m \,|\, \phrase_{\backslash m}, \currimg, \previmg)$ be the text model's predicted probability of token $m$ given $\currimg, \previmg$, and $\phrase_{\backslash m}$ ($\theta$ are the weights of the model). Then, $l(w, p_\theta(\phrase_m \,|\, \phrase_{\backslash m}, \currimg, \previmg))$ is the cross-entropy loss of predicting token $m$ given those inputs.

It is possible for different sentences in a report to refer to the same image finding. Since we mask single tokens at a time, to prevent information leakage from other sentences we consider each sentence in a report independently. Suppose report $\currreport$ consists of $S$ sentences, so we have $\currreport = [\phrase^1, \septoken, \dots, \septoken, \phrase^S]$, where $\phrase^s$ is the tokens of sentence $s$ and $\septoken$ separates sentences.

For a given sample $(\currimg, \previmg, \currreport) \in \multiimageset$ in the test set indexed by $i$, we define 
 \begin{align*}
     \delta_i(m) =  \sum_{s \in S} [ & l(m, p_\theta(\phrase^s_m \,|\, \phrase^s_{\backslash m}, \currimg, \varnothing)) \\
     -& l(m, p_\theta(\phrase^s_m \,|\, \phrase^s_{\backslash m}, \currimg, \previmg)) ]
 \end{align*}
This is the \ac{MLM} loss for predicting $m$ given each \emph{sentence} in the report with and without the prior image. Note that if $m$ does not appear in a given sentence, its contribution to the sum is zero.
The overall $\deltapriorimgloss(m)$ is computed across all samples:
\begin{equation}
\deltapriorimgloss = \frac{1}{N_m} \left(\sum_{i \in \multiimageset^{\mathrm{test}}} \delta_i(m)\right)
\end{equation}
where $N_m$ is the number of \emph{sentences} in reports in $\multiimageset^{\mathrm{test}}$ in which token $m$ appears. This estimate is subject to high variance when $N_m$ is small. Hence, for \Cref{fig:mlm_token-analysis} we filter to tokens $m$ with $N_m \geq 10$.
We collected 931 tokens with $N_m \geq 10$ from the validation set for manual annotation by a board-certified radiologist. The categories, shown in \Cref{fig:mlm_token-analysis} and described in \Cref{tab:semantic_categories} are specific to the radiology domain.

\input{content/figure_tex/fig_cxr-t}

\subsection{Sentence similarity experiment}
\label{sec:sentence_similarity_sup}
The text models are evaluated in isolation to observe if their encoding is sensitive to key clinical observations. To achieve this, we assess the quality of sentence representations obtained from our text model by examining how well the contradiction and paraphrase pairs can be separated in the embedding space. Unlike the traditional NLI task where a model needs to be fine-tuned, here the models are probed in a zero-shot setting and the BERT output token embeddings are utilised. To do so, we encode the sentences from RadNLI and \cxrtbenchmark{}  sentence similarity datasets with the $\clstoken$ token from CXR-BERT-Specialised \cite{boecking2022making} and \temporalbiovil. For PubMedBERT \cite{gu2021domain} and CXR-BERT-General \cite{boecking2022making} which did not directly optimise the $\clstoken$ token during pretraining, we follow \cite{reimers2019sentence} to average the token output embeddings to represent each sentence. 

Cosine similarity is computed between the representations of each sentence pair in the dataset \cite{reimers2019sentence} and is used as logits for the binary classification between paraphrase and contradiction. Note that for RadNLI, we use the subset of `entailment' and `contradiction' pairs and discard the 'neutral' pairs to unify the task across the two datasets. Given the similarities for each sentence pair, we report ROC-AUC and binary-accuracy. For the latter, a threshold value for each method is derived by setting aside a validation set. For this, we perform ten-fold cross validation and tune the threshold with step size of $0.005$ on the validation set.

\subsection{Image registration algorithm}
In \Cref{sec:sota_performance_biovilt}, image registration is applied to pairs of images as a preprocessing step to enable a fair comparison for the baseline approaches (e.g., \biovil \cite{boecking2022making}).
We performed bidirectional multi-scale registration between image pairs optimising an affine transformation (4 degrees of freedom), using \ac{MI} \cite{thevenaz2000optimization_supp} with 128 bins as the similarity criterion.
In more detail, the spatial transformation is characterised by four parameters: two for translation, one for isotropic scaling, and one for rotation.
The optimisation is repeated five times with different random seeds for initialisation, and the run with the highest \ac{MI} is selected to determine the final spatial alignment.
To better identify the correspondences between the scans, bilateral filtering is applied to each image before registration to remove detailed texture whilst preserving edge information \cite{lee2012nonparametric_supp}.
Our implementation is based on the SimpleITK library\cite{lowekamp2013design}.

%% file: content/figure_tex/commands_supp.tex
\newcommand{\plotatt}[4]{\includegraphics[width=\textwidth]{figures/attention_individual/#1_#2_#3_#4}}

\newcommand{\attwidthexp}{0.45\textwidth}
\newcommand{\cxrwidthsupp}{0.47\textwidth}
\newcommand{\nounderscore}[1]{\StrSubstitute{#1}{_}{ }}

%% file: content/figure_tex/table_report_examples.tex
\begin{table*}[ht]
    \centering
    \begin{tabular}{@{} p{0.27\linewidth} p{0.39\linewidth}   p{0.27\linewidth} @{}}
        \toprule
        \textbf{Reference (by radiologist)} & \textbf{\biovil}  & \textbf{\temporalbiovil} \\
        \midrule
         No evidence of acute cardiopulmonary process. \fcolorbox{MidnightBlue}{white}{Decreased} right pleural effusion. & Small right pleural effusion.  & Small right pleural effusion, \fcolorbox{MidnightBlue}{white}{decreased} since \_. \\
         \midrule
        \fcolorbox{MidnightBlue}{white}{No relevant change} as compared to the previous image. No evidence of pneumonia or other parenchymal pathology. Normal size of the heart. No pleural effusions. & \fcolorbox{Sepia}{gray!20}{In comparison} with the study of \_, there has been \fcolorbox{Sepia}{gray!20}{placement} of a right ij port - a - cath that extends to the lower svc. No evidence of post procedure pneumothorax. The cardiac silhouette is within normal limits and there is no vascular congestion or pleural effusion. There is some asymmetry of opacification at the bases, more prominent on the left. In the appropriate clinical setting, this could possibly represent a developing consolidation. & As compared to the previous radiograph, \fcolorbox{MidnightBlue}{white}{no relevant change} is seen. The lung volumes are normal. Normal size of the cardiac silhouette. Normal hilar and mediastinal structures. No pneumonia, no pulmonary edema, no pleural effusions.\\
        \midrule
         \fcolorbox{MidnightBlue}{white}{No previous images}. The cardiac silhouette is within normal limits and there is no evidence of vascular congestion, pleural effusion, or acute focal pneumonia. & \fcolorbox{Sepia}{gray!20}{In comparison} with the study of \_, there is \fcolorbox{Sepia}{gray!20}{little change} and no evidence of acute cardiopulmonary disease. No pneumonia, vascular congestion, or pleural effusion.&
         \fcolorbox{MidnightBlue}{white}{No previous images}. The cardiac silhouette is within normal limits and there is no vascular congestion, pleural effusion, or acute focal pneumonia. 
\\

        \bottomrule
    \end{tabular}
    \caption{
        Comparison between reports generated by radiologists, \biovil using only a single current image and \temporalbiovil using both the current and previous study. \temporalbiovil with access to longitudinal information can generate more accurate reports with more precise details on the progression of findings (as in the first and second example) while avoiding hallucination (in the third example). \fcolorbox{MidnightBlue}{white}{Blue box} highlights the correct temporal information and \fcolorbox{Sepia}{gray!20}{brown box} highlights incorrect temporal information including hallucination.
    }
    \label{tab:nlg}
    \vspace{5 mm}
\end{table*}

%% file: content/figure_tex/fig_cxr-t_loss_label_quality.tex
\begin{figure}[ht]
    \centering
    \includegraphics[width=\linewidth]{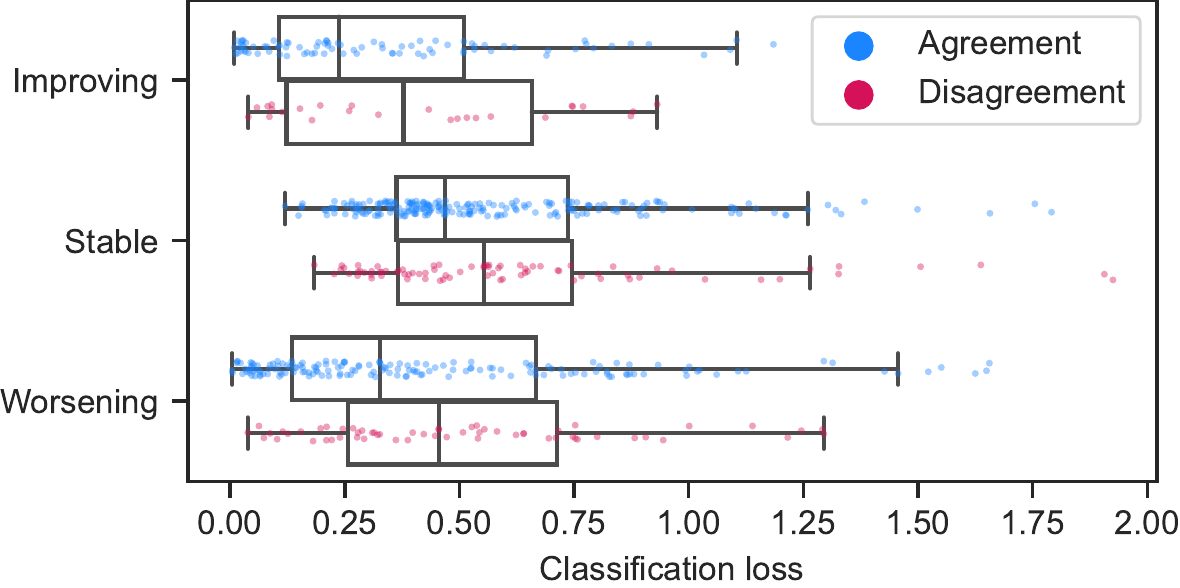}
    \caption{
        Cross entropy between model predictions and \cxrtbenchmark{} temporal classification labels.
        `Disagreement' indicates cases for which annotations differed amongst radiologists.
        Model performance is higher for cases with with low ambiguity (`Agreement').
    }
    \label{fig:cxrtloss}
\end{figure}

%% file: content/figure_tex/fig_attention_supp.tex
\newcommand{\plotprev}[3]{
    \begin{subfigure}{\cxrwidthsupp}
        \centering
        \caption*{Prior image}
        \plotatt{#1}{#2}{#3}{previous}
    \end{subfigure}
}

\newcommand{\plotcurr}[3]{
    \begin{subfigure}{\cxrwidthsupp}
        \centering
        \caption*{Current image}
        \plotatt{#1}{#2}{#3}{current}
    \end{subfigure}
}

\newcommand{\plotcomparethree}[4]{
    \begin{subfigure}{0.31\textwidth}
        \centering
        \plotprev{#1}{#2}{#3}
        \hfill
        \plotcurr{#1}{#2}{#3}
        \caption{Example of #2 \nounderscore{#1}}
        \label{#4}
    \end{subfigure}
}

\begin{figure*}
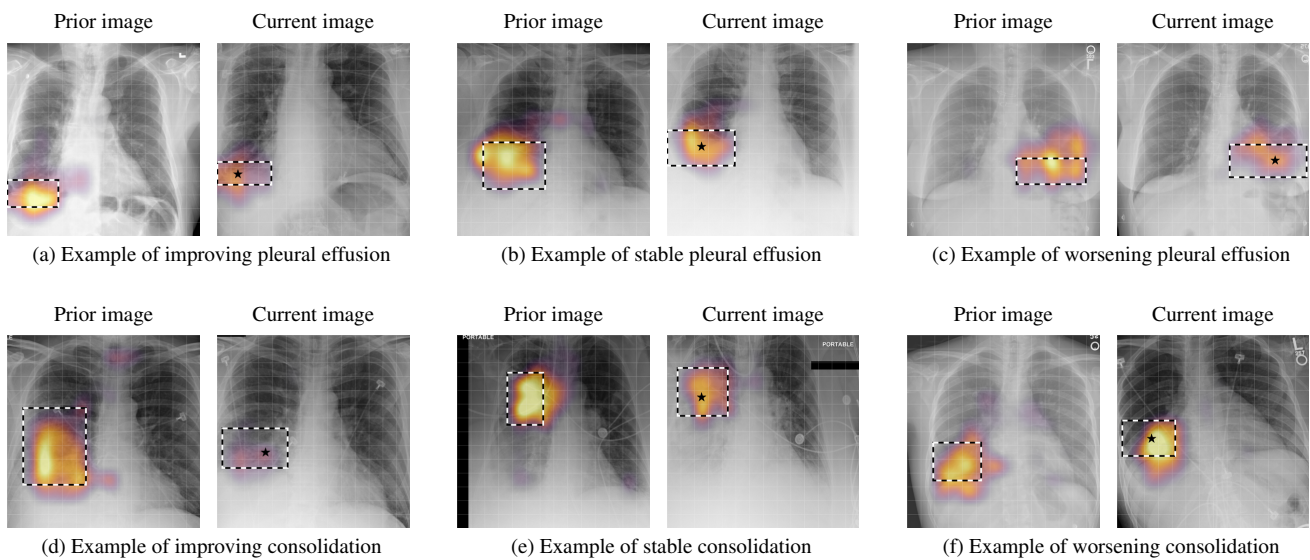

     \centering
     
     \plotcomparethree{pleural_effusion}{improving}{50512608}{fig:effusion_improving}
     \hfill
     \plotcomparethree{pleural_effusion}{stable}{52627981}{fig:effusion_stable}
     \hfill
     \plotcomparethree{pleural_effusion}{worsening}{56535768}{fig:effusion_worsening}
     
     \bigskip
     
     \plotcomparethree{consolidation}{improving}{58145236}{fig:consolidation_improving}
     \hfill
     \plotcomparethree{consolidation}{stable}{54337759}{fig:consolidation_stable}
     \hfill
     \plotcomparethree{consolidation}{worsening}{52445006}{fig:consolidation_worsening}
     
     \caption{
        Self-attention rollout maps~\cite{abnar-zuidema-2020-quantifying} from the reference patch (marked with $\filledstar$) to the current and prior images, overlaid on example cases of \subref{fig:effusion_improving} improving,
        \subref{fig:effusion_stable} stable
        and
        \subref{fig:effusion_worsening} worsening
        pleural effusion (top row) and consolidation (bottom row).
        The bounding boxes, annotated by a radiologist, show the area corresponding to the pathology.
        The centre patch in the bounding box for the current image was selected as reference.
        The grid (14 $\times$ 14) represents the visual tokens processed in the transformer encoder blocks.}
    \label{fig:self_attention_qualitative_examples}

\end{figure*}

\newcommand{\plotprevrot}[4]{
    \begin{subfigure}{\cxrwidthsupp}
        \centering
        \caption*{Prior image}
        \plotatt{#1}{#2}{#3}{previous#4}
    \end{subfigure}
}

\newcommand{\plotcurrrot}[4]{
    \begin{subfigure}{\cxrwidthsupp}
        \centering
        \caption*{Current image}
        \plotatt{#1}{#2}{#3}{current#4}
    \end{subfigure}
}

\newcommand{\plotcomparethreerot}[6]{
    \begin{subfigure}{0.31\textwidth}
        \centering
        \plotprevrot{#1}{#2}{#3}{#6}
        \hfill
        \plotcurrrot{#1}{#2}{#3}{#6}
        \caption{#5}
        \label{#4}
    \end{subfigure}
}

\begin{figure*}
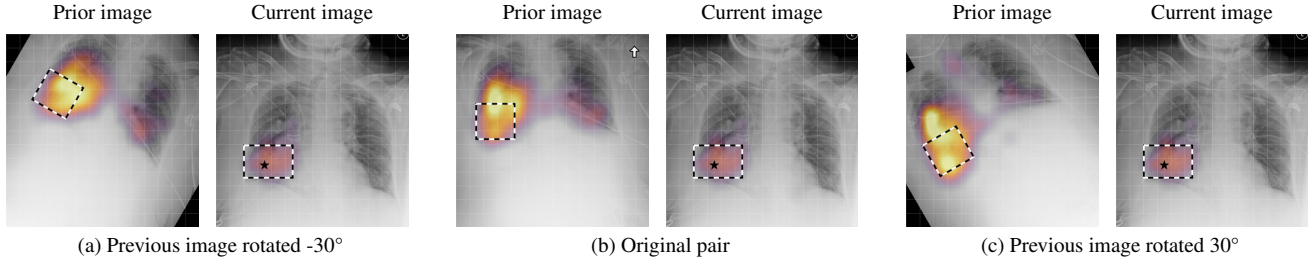

     \centering
     
     \plotcomparethreerot{consolidation}{stable}{51139759}{fig:att_-30_deg}{Previous image rotated -30\degree}{_rotated_-30}
     \hfill
     \plotcomparethreerot{consolidation}{stable}{51139759}{fig:att_0_deg}{Original pair}{}
     \hfill
     \plotcomparethreerot{consolidation}{stable}{51139759}{fig:att_30_deg}{Previous image rotated 30\degree}{_rotated_30}
     
     \caption{
        Comparison of roll-out maps computed after applying in-plane spatial rotations to the prior image.
        The reference visual token ($\filledstar$) attends to the corresponding anatomical region annotated by an expert independent of the underlying spatial transformation.
    }
    \label{fig:self_attention_pose_variations_supp}
\end{figure*}

%% file: content/figure_tex/fig_data_curation.tex
\begin{figure*}
    \centering
    \begin{subfigure}[b]{0.195\linewidth}
        \centering
        \begin{subfigure}[b]{0.50\linewidth}
            \centering
            \includegraphics[height=1.85cm]{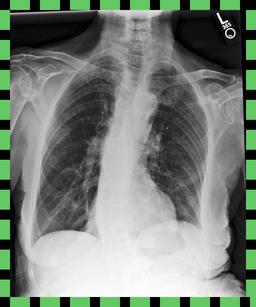}
        \end{subfigure}%
        \hfill
        \begin{subfigure}[b]{0.50\linewidth}
            \centering
            \includegraphics[height=1.85cm]{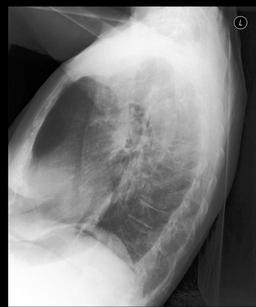}
        \end{subfigure}
        \subcaption{Incorrect view}
        \label{fig:auto_qc_view}
    \end{subfigure}
    \begin{subfigure}[b]{0.195\linewidth}
        \centering
        \begin{subfigure}[b]{0.50\linewidth}
            \centering
            \includegraphics[height=1.85cm]{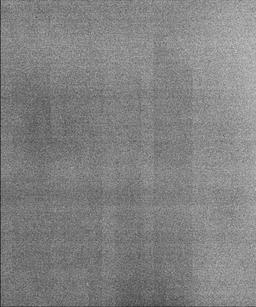}
        \end{subfigure}%
        \hfill
        \begin{subfigure}[b]{0.50\linewidth}
            \centering
            \includegraphics[height=1.85cm]{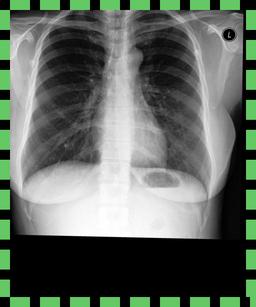}
        \end{subfigure}
        \subcaption{Invalid acquisition}
        \label{fig:auto_qc_noise}
    \end{subfigure}
    \begin{subfigure}[b]{0.195\linewidth}
        \centering
        \begin{subfigure}[b]{0.50\linewidth}
            \centering
            \includegraphics[height=1.85cm]{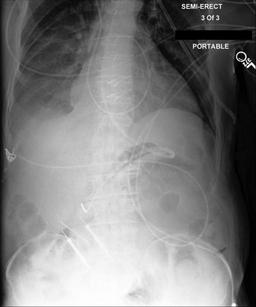}
        \end{subfigure}%
        \hfill
        \begin{subfigure}[b]{0.50\linewidth}
            \centering
            \includegraphics[height=1.85cm]{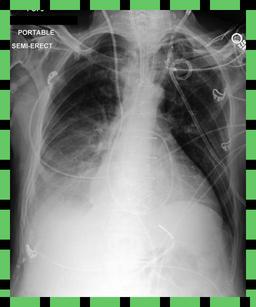}
        \end{subfigure}
        \subcaption{Incomplete field of view}
        \label{fig:auto_qc_fov}
    \end{subfigure}
    \begin{subfigure}[b]{0.195\linewidth}
        \centering
        \begin{subfigure}[b]{0.50\linewidth}
            \centering
            \includegraphics[height=1.85cm]{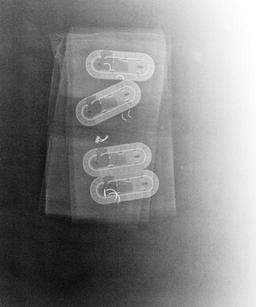}
        \end{subfigure}%
        \hfill
        \begin{subfigure}[b]{0.50\linewidth}
            \centering
            \includegraphics[height=1.85cm]{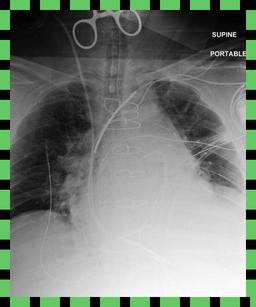}
        \end{subfigure}
        \subcaption{Non-human sample}
        \label{fig:auto_qc_object_1}
    \end{subfigure}
    \begin{subfigure}[b]{0.195\linewidth}
        \centering
        \begin{subfigure}[b]{0.50\linewidth}
            \centering
            \includegraphics[height=1.85cm]{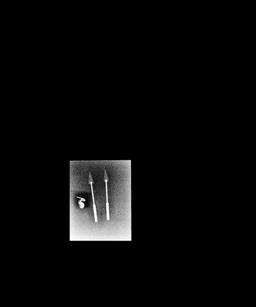}
        \end{subfigure}%
        \hfill
        \begin{subfigure}[b]{0.50\linewidth}
            \centering
            \includegraphics[height=1.85cm]{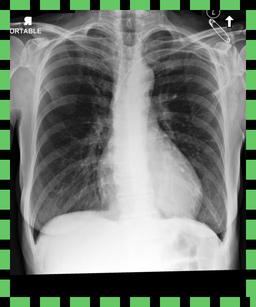}
        \end{subfigure}
        \subcaption{Non-human sample}
        \label{fig:auto_qc_object_2}
    \end{subfigure}

    \bigskip

    \begin{subfigure}[b]{0.195\linewidth}
        \centering
        \begin{subfigure}[b]{0.50\linewidth}
            \centering
            \includegraphics[height=1.85cm]{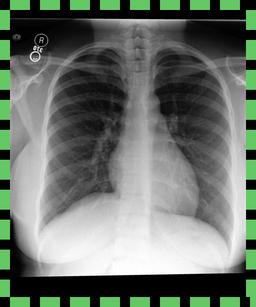}
        \end{subfigure}%
        \hfill
        \begin{subfigure}[b]{0.50\linewidth}
            \centering
            \includegraphics[height=1.85cm]{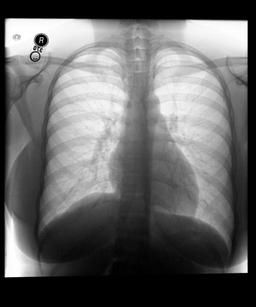}
        \end{subfigure}
        \subcaption{Inverted intensities}
        \label{fig:auto_qc_inverted}
    \end{subfigure}
    \begin{subfigure}[b]{0.195\linewidth}
        \centering
        \begin{subfigure}[b]{0.50\linewidth}
            \centering
            \includegraphics[height=1.85cm]{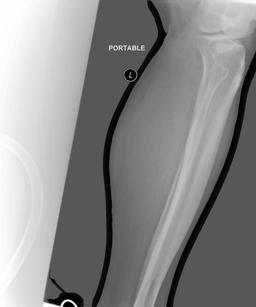}
        \end{subfigure}%
        \hfill
        \begin{subfigure}[b]{0.50\linewidth}
            \centering
            \includegraphics[height=1.85cm]{figures/auto_qc/p16599161_s51926993_84dddb4e-0a4caa35-216da517-31ab25e3-b2a8361e}
        \end{subfigure}
        \subcaption{Non-chest sample}
        \label{fig:auto_qc_leg}
    \end{subfigure}
    \begin{subfigure}[b]{0.195\linewidth}
        \centering
        \begin{subfigure}[b]{0.50\linewidth}
            \centering
            \includegraphics[height=1.85cm]{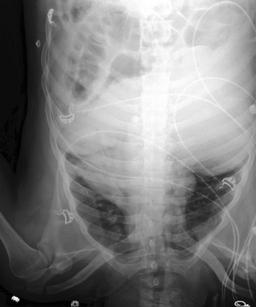}
        \end{subfigure}%
        \hfill
        \begin{subfigure}[b]{0.50\linewidth}
            \centering
            \includegraphics[height=1.85cm]{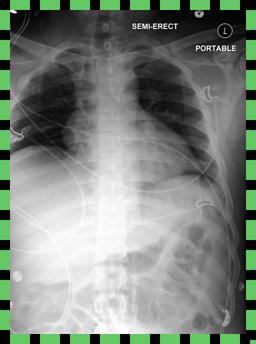}
        \end{subfigure}
        \subcaption{Image orientation}
        \label{fig:auto_qc_upside_down}
    \end{subfigure}
    \begin{subfigure}[b]{0.195\linewidth}
        \centering
        \begin{subfigure}[b]{0.50\linewidth}
            \centering
            \includegraphics[height=1.85cm]{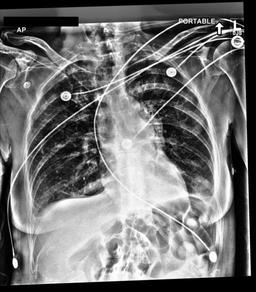}
        \end{subfigure}%
        \hfill
        \begin{subfigure}[b]{0.50\linewidth}
            \centering
            \includegraphics[height=1.85cm]{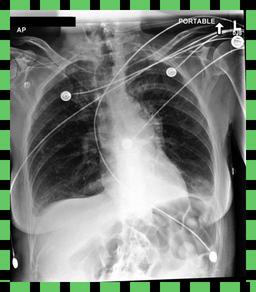}
        \end{subfigure}
        \subcaption{Post-processed image}
        \label{fig:auto_qc_clahe}
    \end{subfigure}
    \begin{subfigure}[b]{0.195\linewidth}
        \centering
        \begin{subfigure}[b]{0.50\linewidth}
            \centering
            \includegraphics[height=1.85cm]{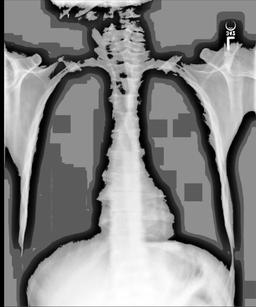}
        \end{subfigure}%
        \hfill
        \begin{subfigure}[b]{0.50\linewidth}
            \centering
            \includegraphics[height=1.85cm]{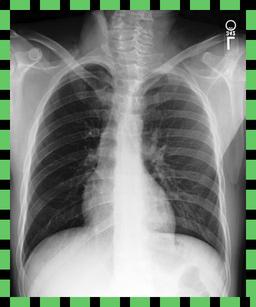}
        \end{subfigure}
        \subcaption{Processing artefacts}
        \label{fig:auto_qc_artifacts}
    \end{subfigure}
    
    \caption{
        Pairwise ranking of images performed by the proposed data curation method (see \Cref{sec:adaptations_to_downstream}) on images from the MIMIC-CXR v2 dataset.
        Images highlighted with dashed green rectangles are automatically selected by our method and used for training to improve model's downstream performance.
        The rejected image samples may not be appropriate for training due to image acquisition or image processing issues as shown in each subfigure above.
    }
    \label{fig:data_curation_supp}
\end{figure*}

%% file: content/figure_tex/fig_cxr-t.tex
\newcommand{\plotcxrt}[1]{\includegraphics[width=\textwidth]{figures/cxr-t/#1}}

\newcommand{\plotprevcxrt}[1]{
    \begin{subfigure}{\cxrwidthsupp}
        \centering
        \caption*{Prior image}
        \plotcxrt{#1_previous}
    \end{subfigure}
}

\newcommand{\plotcurrcxrt}[1]{
    \begin{subfigure}{\cxrwidthsupp}
        \centering
        \caption*{Current image}
        \plotcxrt{#1_current}
    \end{subfigure}
}

\newcommand{\plotcomparethreecxrt}[4]{
    \begin{subfigure}{0.31\textwidth}
        \centering
        \plotprevcxrt{#1_ensemble_correct_certain_#3_#4}
        \hfill
        \plotcurrcxrt{#1_ensemble_correct_certain_#3_#4}
        \caption{\capitalisewords{#3} #2}
        \label{fig:cxrt_#1_#3}
    \end{subfigure}
}

\begin{figure*}
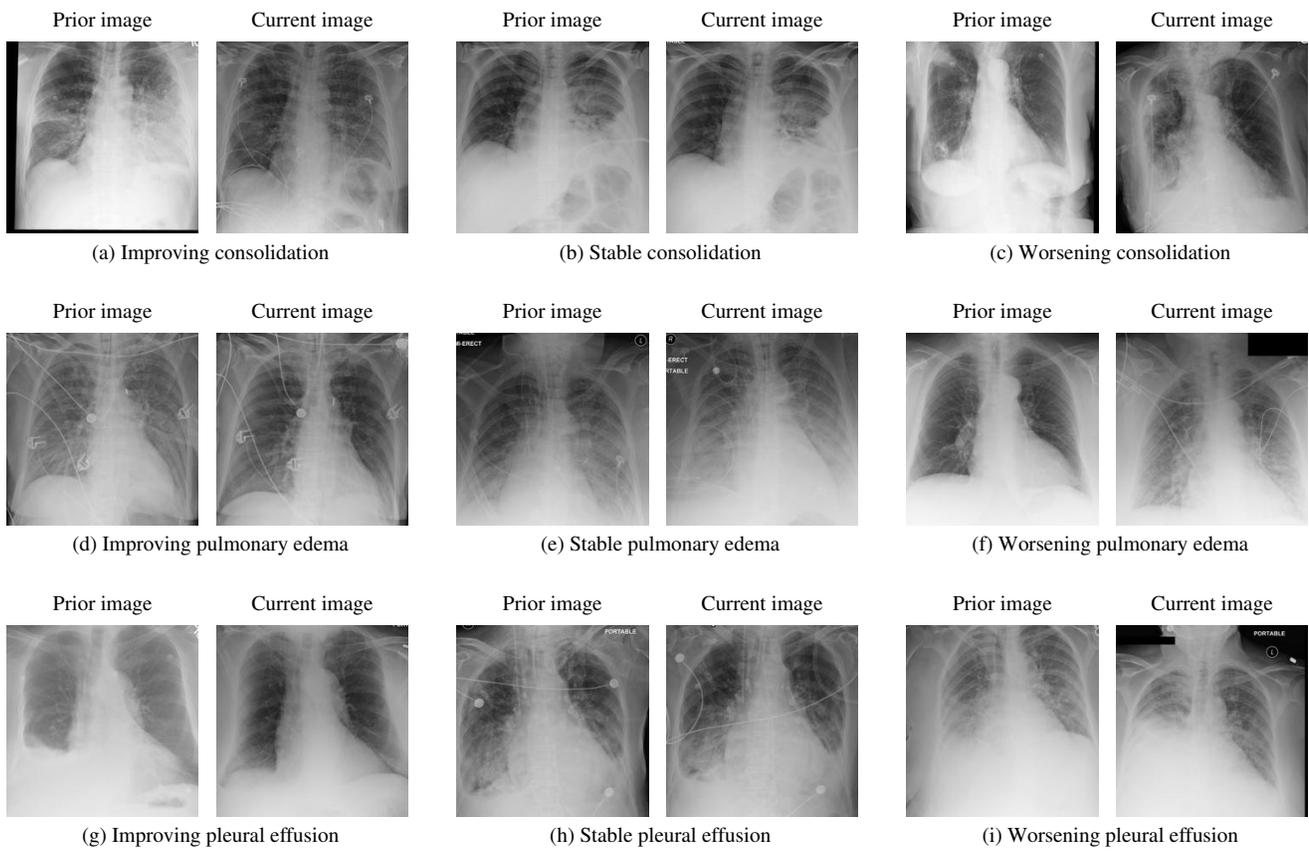

     \centering
     \plotcomparethreecxrt{consolidation}{consolidation}{improving}{k_0_idx_24_study_58856479_quality_multiple_experts}
     \hfill
     \plotcomparethreecxrt{consolidation}{consolidation}{stable}{k_4_idx_20_study_50414777_quality_multiple_experts}
     \hfill
     \plotcomparethreecxrt{consolidation}{consolidation}{worsening}{k_0_idx_18_study_56785501_quality_multiple_experts}
     
     \par\bigskip
     
     \plotcomparethreecxrt{edema}{pulmonary edema}{improving}{k_1_idx_111_study_51310486_quality_one_expert}
     \hfill
     \plotcomparethreecxrt{edema}{pulmonary edema}{stable}{k_2_idx_46_study_58240924_quality_one_expert}
     \hfill
     \plotcomparethreecxrt{edema}{pulmonary edema}{worsening}{k_0_idx_142_study_56836467_quality_one_expert}
     
     \par\bigskip
     
     \plotcomparethreecxrt{pleural_effusion}{pleural effusion}{improving}{k_0_idx_122_study_57015680_quality_multiple_experts}
     \hfill
     \plotcomparethreecxrt{pleural_effusion}{pleural effusion}{stable}{k_0_idx_76_study_50713452_quality_multiple_experts}
     \hfill
     \plotcomparethreecxrt{pleural_effusion}{pleural effusion}{worsening}{k_05_idx_159_study_55148162_quality_multiple_experts}
          
     \caption{
        Examples of image pairs in our \cxrtbenchmark\ benchmark.
    }
    \label{fig:cxrtimgs}
\end{figure*}